\documentclass[11pt]{article}

\usepackage[final]{acl}
\usepackage{amsmath}
\usepackage{amssymb}
\usepackage{times}
\usepackage{latexsym}
\usepackage{booktabs}
\usepackage{multirow}
\usepackage[table]{xcolor}
\usepackage[T1]{fontenc}
\usepackage[utf8]{inputenc}

\usepackage{microtype}

\usepackage{inconsolata}

\usepackage{graphicx}
\graphicspath{{../paper/figures/}{../paper/figures/reviewed/}{v2_short/figures/}{figures/}}

\title{Where Do Apparent LLM Clinical Triage Failures Arise? Localizing the Multiple-Choice Format Effect}

\author{
  \textbf{David Fraile Navarro\textsuperscript{1}},
  \textbf{Berardino Como\textsuperscript{2}},
  \textbf{Jialei Sheng\textsuperscript{3}},
  \textbf{Soundariya Ananthan\textsuperscript{4}},
  \textbf{Shlomo Berkovsky\textsuperscript{1}}
\\[3pt]
  \textsuperscript{1}Macquarie University, Sydney, Australia \quad
  \textsuperscript{2}Politecnico di Bari, Bari, Italy
\\
  \textsuperscript{3}NSW Health, Sydney, Australia \quad
  \textsuperscript{4}Independent Researcher
\\[3pt]
  \small{\texttt{david.frailenavarro@mq.edu.au}, \texttt{b.como@studenti.poliba.it}, \texttt{jialei.sheng@health.nsw.gov.au},}
\\
  \small{\texttt{a.soundariya@gmail.com}, \texttt{shlomo.berkovsky@mq.edu.au}}
\\[3pt]
  \small{\textbf{Correspondence:} \href{mailto:david.frailenavarro@mq.edu.au}{david.frailenavarro@mq.edu.au}}
}
\begin{document}
\maketitle
\begin{abstract}
LLM evaluations using clinician-authored triage vignettes have reported
substantial under-triage under constrained multiple-choice testing.
Yet model performance on the same clinical cases can change when
responses are generated in free text. We test whether this format
effect appears while the case is processed or when clinical information
is mapped to the final answer. Using sparse-autoencoder (SAE) features in Gemma 3 4B/12B
IT and Qwen3-8B, we find that medical features fire on the shared
clinical narrative under both formats but are inactive at the
multiple-choice decision token. Emergency-tier information is
linearly decodable from vignette representations with ROC-AUC
$0.95$--$1.00$ under both formats, with no significant format
difference, but is attenuated at the decision token. Natural-language
autoencoder verbalization and top-feature characterization associate
that token with the multiple-choice scaffold. In a direct linear
projection, the identified medical features contribute zero, whereas
scaffold-peaking features account for over $91\%$ of unsigned
attribution in both Gemma models. Behaviorally, whether multiple
choice improves or worsens performance depends on the model.
Option-order shuffles rule out simple positional bias, and cases that
differ between formats are usually one severity tier apart. Together,
these findings place the strongest correlates of the format effect at
answer selection while leaving open whether unmeasured clinical
representations also differ.

Code and data to reproduce experiments are available in the study repository\footnote{\url{https://github.com/dafraile/SAE_mad}}.
\end{abstract}

\section{Introduction}

Large language models (LLMs) can perform strongly on medical
questions~\cite{singhal2023large}, but a structured clinical-triage
benchmark using clinician-authored vignettes reported a $51.6\%$
under-triage rate for emergency cases~\cite{ramaswamy2026chatgpt}.
A behavioral replication of the same clinical scenarios found
substantially different model performance when structured,
forced-letter testing was compared with naturalistic, free-text
interaction~\cite{frailenavarro2026triage}, consistent with broader
evidence of prompt-format and option-order sensitivity
~\cite{zheng2024large,pezeshkpour2024large,sclar2024quantifying}.
These findings suggest that the measured failure rate is partly
contingent on evaluation format, but behavioral results alone cannot
show where in the model the format-dependent difference arises.

This raises a question that behavior alone cannot
answer: \emph{where in the model do correlates of the format effect appear?} Two
hypotheses are plausible.
\textbf{Representation-shift hypothesis}: the format effect arises
before answer selection. Appending the multiple-choice scaffold changes
how the model represents the clinical case relative to free-text generation.
\textbf{Output-mapping hypothesis}: prompt format changes how the model
maps an already-formed clinical representation onto the final answer;
the clinical representation remains similar, but the format affects
how that information is translated into the model's response.

This distinction has well-established conceptual
precedent~\cite{burns2023discovering,kadavath2022language,turpin2023language}.
Concurrent work in clinical triage specifically documents a 53-pp
knowledge--action gap on Qwen 2.5 7B Instruct, with four mechanistic
interventions, including SAE feature steering, failing to reliably
correct it~\cite{basu2026interpretability}. We ask a different,
more specific question: what differs when output format is varied
while the clinical-content portion of the prompt is held unchanged?

\textbf{Contributions.} We combine SAE features, natural-language
autoencoders, linear probes, and behavioral decomposition across
three instruction-tuned LLMs from two families to localize
correlates of the format effect:

\begin{itemize}
\setlength\itemsep{0pt}
\item Identified medical features peak on the clinical narrative
under both output formats. Separate acuity probes decode emergency
status at ROC-AUC $0.95$--$1.00$ from that vignette representation,
with no significant format difference.

\item At the multiple-choice decision token, those medical
features are inactive and urgency decodability is reduced. NLA and
top-feature analysis associate this state with the scaffold; an exact
linear projection assigns over $91\%$ of unsigned attribution to
scaffold-peaking features and zero to the identified medical features.

\item We decompose the performance differences between multiple-choice
and free-text into shifts between triage levels and response deferral.
Most gap-driving differences are between nearby triage levels, while
deferral is primarily a separate label-space issue. Exhaustive
option-order shuffles further distinguish preferences for particular
triage recommendations from simple positional bias.

\item Linear probes predict case-level correctness
flips between formats from the source-format hidden state alone.
Moving the formatting constraint to the beginning of the prompt
substantially changes this predictive signal, showing that it is
sensitive to where the constraint appears in the prompt.

\end{itemize}

Across models, the strongest measured correlates of the format effect
appear near answer selection, while the identified medical features
respond similarly across formats during vignette processing and urgency
is similarly decodable. This supports an output-mapping account for the
measured signals without ruling out other representational differences.

\begin{figure*}[t]
  \centering
  \includegraphics[width=\textwidth]{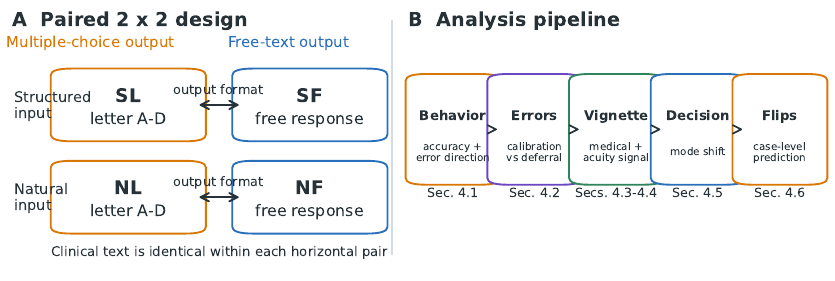}
  \caption{Study overview. Within each input style, the clinical
  text is held fixed while output format changes (left). The
  analyses progressively test behavior, error composition,
  vignette-level representation, decision-token localization, and
  case-level predictability (right).}
  \label{fig:overview}
\end{figure*}

\section{Background and Related Work}

Emergency triage is an initial prioritization under time pressure and limited
information, distinct from final diagnosis~\cite{yancey2023emergency}.
For this benchmark, the target is therefore an acuity disposition
from sparse patient-reported information, not a diagnosis.

Med-PaLM established strong medical question-answering performance
~\cite{singhal2023large,singhal2025expert}, while structured
clinical-triage evaluations have exposed clinically important failures
~\cite{ramaswamy2026chatgpt} in recent LLMs. Concurrent mechanistic work on this
task reports a $53$-pp knowledge--action gap on Qwen 2.5 7B
Instruct and finds that four interventions do not reliably correct
it~\cite{basu2026interpretability}. We instead hold the clinical text
fixed and vary output format to ask where the resulting differences
appear in the model.

Medical MCQs can also \emph{inflate} apparent competence: models
solve fictional medical questions above physicians by exploiting
shallow cues~\cite{griot2025pattern}. Our results show that both
directions can occur. Multiple-choice helps at 12B and Qwen but hurts at 4B,
where shuffling the letter--content mapping recovers free-text-level
accuracy. Our prior frontier-model study likewise finds a penalty
for GPT-5.2 and two Gemini models but not two Claude models in a
matched emergency-asthma ablation (forced $0$--$24\%$ vs. free-text
$100\%$, $n{=}25$ per cell; all $p{<}10^{-8}$), and a smaller
aggregate naturalistic advantage ($70.1\%$ vs. $63.6\%$,
$p{=}0.015$)~\cite{frailenavarro2026triage}. Thus the sign and size
of the format effect are model-dependent rather than monotonic in
scale; our case study asks where its correlates appear in open
models with released SAEs.

\section{Methods}

\subsection{Dataset}
\label{sec:dataset}
We use all 60 clinical vignettes from the replication corpus
of~\cite{frailenavarro2026triage}, reconstructed from the
clinical-triage benchmark of~\cite{ramaswamy2026chatgpt}. The
corpus varies both input style (structured vignettes or
naturalistic/conversational symptom descriptions) and output format
(multiple-choice or free-text) in a $2 \times 2$
factorial, yielding four conditions: \textbf{SL} (structured +
multiple-choice), \textbf{NL} (natural + multiple-choice),
\textbf{SF} (structured + free-text), and \textbf{NF} (natural +
free-text). In our condition codes the first letter denotes
input (\textbf{S}tructured or \textbf{N}atural), and the second
letter denotes output (\textbf{L}etter for forced-letter
multiple-choice, \textbf{F} for \textbf{f}ree-text). Throughout,
`multiple-choice' refers to prompts that select a 4-option A-D 
single-letter response; we call this appended instruction block the \textbf{multiple-choice scaffold}. Within each input format the forced-letter and
free-text conditions are byte-identical in clinical content
(NL-NF at 1033 characters/case) and differ only in whether the
multiple-choice scaffold ($50$--$60$ tokens) is appended. NL-NF
and SL--SF therefore each isolate output format at a fixed input.
Mechanistic analyses use the NL-NF pair throughout; SL-SF appears
as a behavioral robustness check in
Section~\ref{sec:phase0_5_cells} and as an input-style mechanistic
robustness check in Appendix~\ref{app:input-style-robustness}. Output gold labels follow the
source benchmark's four care tiers (A: monitor at home; B: see
a doctor in a few weeks; C: see a doctor in 24-48 hours; D: go
to emergency). Half the cases have two acceptable gold labels
(e.g., $C/D$ in $24/60$ cases), and either is scored as correct
(full distribution in Appendix~\ref{app:dataset}).

\subsection{Models and SAEs}
\label{sec:models-saes}

Both SAE pipelines decompose residual streams into sparse
interpretable features~\cite{bricken2023monosemanticity,cunningham2023sparse,templeton2024scaling}:
Gemma Scope 2 uses JumpReLU~\cite{rajamanoharan2024jumping,lieberum2024gemma},
while Qwen-Scope uses TopK with $k{=}100$~\cite{makhzani2013ksparse,gao2024scaling,qwen2026scope}.
We use Gemma 3 4B IT (34 layers, $d_\text{model}=2560$) and Gemma 3
12B IT (48 layers, $d_\text{model}=3584$) with Gemma Scope 2 SAEs
($d_\text{SAE}=16{,}384$, $\sim$$14\%$ reconstruction error), and
Qwen3-8B (36 layers, $d_\text{model}=4096$) with Qwen-Scope SAEs
($d_\text{SAE}=65{,}536$, $\sim$$34\%$ reconstruction error at L31).
For Gemma we sweep four matched-depth layers at
$\approx 27\%, 50\%, 65\%, 85\%$ of total depth: in 4B layers 9, 17,
22, 29 and in 12B layers 12, 24, 31, 41. For Qwen we use
layer 31 at $\approx 86\%$ relative depth, selected by
reconstruction error from a four-layer pilot sweep
(Appendix~\ref{app:qwen-layer-selection}).

\subsection{Behavioral test}
\label{sec:scoring}

For each case we generate output via greedy decoding. SL and NL
answers are extracted via \texttt{regex}. NF and SF free-text are scored by \texttt{gpt-5.2-thinking-high} and
\texttt{claude-sonnet-4.6} LLM judges, adjudicating against the gold label
on the A--D scale, with acceptable matches on
dual-labeled cases. Letter-extraction (SL, NL) and LLM-judge (SF, NF) pipelines therefore score slightly different artifacts (the emitted letter vs.\ an interpreted commitment from free text); we treat this as
part of the format contrast, not a noise source to be removed. We
report per-condition accuracy, and
inter-rater agreement and Cohen's $\kappa$ as a calibration signal.
A clinician-adjudicated subset (Section~\ref{sec:deferral-adjudication}, Appendix~\ref{app:clinician}) provides external calibration of the LLM judges against clinical judgment on a stratified 16-case sample.

%\subsection{Option-order shuffle}
\label{sec:option-order}
To disambiguate whether multiple-choice accuracy depends on letter position or letter content, we
run an exhaustive shuffle of the letter$\to$content mapping in
NL. For each of the $60$ NL prompts, we generate all $23$ non-identity permutations of the assignment $\{A, B, C, D\}$, keeping
option texts verbatim but rewriting the letter labels 
(e.g., ``see a doctor in 24-48 hours'' may sit under any letter). 
We re-run multiple-choice generation on each shuffled prompt under greedy decoding
($60 \times 23 = 1380$ NL passes per model). For each
shuffle we record both the picked letter and the picked content. Position bias predicts a
constant picked letter across shuffles; a content prior predicts
constant picked content. Case-clustered $95\%$ CIs are
bootstrapped over cases ($2{,}000$ resamples; full protocol in
Appendix~\ref{app:option_shuffle}).

\subsection{Mechanistic invariance test}
\label{sec:mech-invariance}

\paragraph{Notation.}
For prompt $p$ with $T_p$ tokens, $a_f(t,p)$ denotes SAE feature
$f$'s activation at token $t$. Its max-pooled prompt activation is
\begin{equation}
a_f(p) = \max_{1\leq t\leq T_p} a_f(t,p).
\label{eq:maxpool}
\end{equation}
For each (model, layer, case, condition) we encode the residual
stream through the SAE and apply Eq.~\ref{eq:maxpool} over
user-content tokens (pooling conventions in
Appendix~\ref{app:pooling-conventions}). Max-pooling avoids dilution
under Qwen-Scope's $k{=}100$ TopK sparsity; Gemma aggregation checks
are in Appendix~\ref{app:metric_consistency}.
We compare two feature subsets $\mathcal{F}$: the $3$
contrastively-identified medical features per layer
(Section~\ref{sec:medical-feature-id}) and a magnitude-matched
\emph{random} control of $30$ features drawn per layer from the
non-medical feature pool (size $500$--$2{,}200$; full protocol in
Appendix~\ref{app:why-three-features}). For each case and
each subset $\mathcal{F}$, we
summarize NL-NF invariance by two statistics. The first is the
symmetric mean absolute percentage error
(sMAPE,~\citealp{makridakis1993accuracy}) in its standard form,
applied here over the features in $\mathcal{F}$:
\[
\mathrm{sMAPE}_{\mathcal{F}} \;=\; \frac{1}{|\mathcal{F}|}
\sum_{f \in \mathcal{F}}
\frac{|a^f_{\mathrm{NL}} - a^f_{\mathrm{NF}}|}
     {(|a^f_{\mathrm{NL}}| + |a^f_{\mathrm{NF}}|)/2},
\]
where $a^f_{\mathrm{NL}}$ and $a^f_{\mathrm{NF}}$ are feature $f$'s
max-pooled activations in the NL and NF conditions; lower $=$ more
invariant. In implementation we floor the denominator at
$\varepsilon=10^{-8}$ to prevent division by zero when both
activations are exactly zero; this affects only inactive features
and leaves active-feature scores unchanged. The second statistic is
cosine similarity between the NL and NF feature-subset activation
vectors (higher $=$ more invariant).
%; effective $n$ is reported when either vector is zero).

We stratify case-level results by joint NL-NF correctness into
five strata: both-right, both-wrong, NF-only-right (NL wrong, NF
correct), NL-only-right (the inverse), and
judges-disagree (withheld from headline tables). For each (layer,
stratum) we report bootstrap $95\%$ CIs on the per-case
medical-random $\Delta$ for both sMAPE and cosine. The
bootstrap resamples the $n$ cases in the stratum with replacement,
computes the resample mean, repeats this $2{,}000$ times, and
takes the $2.5$\textsuperscript{th} and
$97.5$\textsuperscript{th} percentiles.

\paragraph{Analyzed late layer.}
We report one representative late layer per model in the main text
(4B L29, 12B L31, Qwen L31), with the full Gemma layer sweeps and
Qwen pilot in Appendix~\ref{app:full-tables}. At Gemma 12B, the
medical--random difference reverses between shallow/mid layers
(L12, L24) and late layers (L31, L41); our conclusions are therefore
restricted to the late-layer regime, where the pattern is consistent.

\paragraph{Acuity-specific probe.}
To test whether urgency is similarly decodable across formats beyond
the broad medical-domain features, we probe whether the gold tier includes emergency care
($D$; $28$ positive/$32$ negative cases). At each model's headline
late layer, L2 logistic regression ($C{=}0.05$, liblinear, balanced
weights) uses SAE activations max-pooled over vignette tokens or
taken at the decision token, under NL and NF separately. We report
leave-one-out ROC-AUC, empirical $p$ from $1{,}000$ label
permutations with full refitting, and case-bootstrap $95\%$ CIs on
paired AUC differences. Four-class most-urgent-tier macro-OVR AUC
is a secondary check.

\subsection{Decision-token analyses}
\label{sec:methods_decision_token}

We target the hidden state at the NL pre-generation token (the
last user-message token, whose hidden state drives the first
generated letter) with one primary analysis and two SAE-pool
corroborations.

%\paragraph{NLA verbalization (primary).}
On Gemma 3 12B IT (the only open-weight NLA checkpoint we have
access to~\cite{frasertaliente2026nla}), at layer $32$, we capture
residual-stream activations at seven token positions
(\textsc{content}, \textsc{decision}, and four
\texttt{letter\_A/B/C/D}; full positions in
Appendix~\ref{app:nla-details}), pass them through the released
NLA, and have the two LLM judges classify each generated
description on two axes: (\textsc{medical} and \textsc{scaffold},
each \textsc{primary}/\textsc{partial}/\textsc{no}).

%\paragraph{SAE-pool corroborations.}
Two complementary analyses use the SAE feature pool at the same
decision token. Let $W_\mathrm{dec}[:,f]$ be feature $f$'s decoder
direction, $W_U[:,t_\ell]$ the unembedding direction for the token
of letter $\ell\in\{A,B,C,D\}$, and $a_f$ the decision-token
activation. The feature's direct linear contribution is
\begin{equation}
\mathrm{contrib}(f,\ell)=
a_f\,W_\mathrm{dec}[:,f]^\top W_U[:,t_\ell].
\label{eq:contrib}
\end{equation}
A positive value means that feature $f$ raises letter $\ell$'s
logit in this direct projection. For each case, we classify every
active non-medical feature by whether its peak occurs in the shared
vignette or the appended scaffold, then sum absolute contributions
without within-category cancellation. We also characterize the
top-$20$ features by activation and compare NL/NF overlap. Full
methodology, bootstrap intervals, and linear-projection caveats are in
Appendices~\ref{app:logit_attribution}
and~\ref{app:decision_token_features}.

\subsection{Predicting behavioral instability from source-format
representation}
\label{sec:methods_probe}

To test whether source-format hidden states predict which cases change
correctness after a format change, we train L2-regularized
logistic-regression probes
on Gemma 3 4B IT last-token hidden states ($d{=}2,560$) 
at layers $\{9,17,19,22,25,27,29,31,33,34\}$. For
each format transition (e.g., NL$\to$SL), the binary target
is whether correctness flips. The last-token embedding is the only
representation that has attended to
the full input under the model's causal mask, making it the most
direct readout of pre-generation state. Under the
\emph{Constraint-First} (CF) condition, we move the formatting
instructions to the start of the prompt to test whether the
predictive signal depends on the constraint's position.
Training uses nested leave-one-out cross-validation: each outer
fold holds out one case, the inner loop selects regularization on
the remaining $59$, and all fits use balanced class weights. We
report ROC-AUC and PR-AUC with $p$-values from $1{,}000$-iteration
permutation tests with full refitting
per layer and transition. ROC-AUC reflects ranking ability across
the sample, while PR-AUC is interpreted relative to positive-class
prevalence, the no-skill baseline.
%Both metrics are reported to distinguish robust signal from inflated rank-order performance. 

\subsection{Direction-of-format-effect test}
\label{sec:direction-test}

To localize the NL-NF residual-stream
direction in the SAE basis, we average per-case residual
differences into $\Delta r$ and rank SAE encoder columns by their
absolute cosine alignment with this direction,
$|\cos(\Delta r,\,W_\text{enc}[:,f])|$, where $W_\text{enc}[:,f]$
is the encoder column for feature $f$ (i.e.\ the direction the
SAE uses to detect feature $f$ in the residual stream). We then
report medical-feature percentile ranks in this ordering ($0\%$
is top-aligned). Per-case alignment with the aggregate direction is
reported in Appendix~\ref{app:direction-consistency}. NL is longer than NF by the appended
multiple-choice scaffold ($50$--$60$ tokens), so different
aggregation choices conflate the format direction with token
counts. We therefore run three aggregations to separate scaffold
length from format: \textbf{full mean-pool} averages
per-token residuals over each condition's full content range (so
NL includes the scaffold); \textbf{length-controlled
mean-pool} averages only over the byte-identical clinical-prefix
tokens that both conditions share (no scaffold contribution);
\textbf{max-pool} takes each feature's peak activation over
content tokens, avoiding the token-count dilution introduced by
mean-pooling.

\subsection{Deferral adjudication}
\label{sec:deferral-adjudication}

Headline accuracies are 4-way throughout; the 5-way pass never
alters them. The 4-way scoring forces judges to convert each free-text
response to one letter.%, including responses that explicitly decline
%to commit. 
Inspection of 12B NF outputs revealed \emph{deferrals}:
that condition on missing clinical information, %and asks the patient for it 
(e.g., ``ER if BP$>$180/120;
same-day clinic if 140--180; review within two weeks otherwise''). We re-adjudicate every NF response under
$\{A, B, C, D, \text{\textsc{deferred}}\}$ with the same two LLM
judges (full protocol, label definition, and stratum-mapping rules in
Appendix~\ref{app:deferred_adjudication}). To check if the
deferrals are clinically appropriate, a clinician co-author (J.S.),
blinded to LLM-judge labels and model identity, adjudicated a
16-case sample. The sample is stratified across four cells of $4$ cases each:
three at 12B (12B-both-deferred, 12B-clear-correct,
12B-clear-incorrect, covering the deferral phenomenon at the
scale where it occurs) plus one cross-scale cell at 4B
(4B-format-flipped, NL letter wrong but NF judged correct, our
gap-driver stratum). Per-case
appropriateness judgments and the
clinician's deferral rate are reported in
Appendix~\ref{app:clinician}.

\subsection{Medical-feature identification}
\label{sec:medical-feature-id}

For Gemma 3 4B IT we use $3$ medical features per analyzed layer,
identified in~\cite{frailenavarro2026saemad} via a $6$-condition cross-lingual contrastive over MMLU question stems (three languages
$\times$ medical/non-medical; full protocol in
Appendix~\ref{app:medical-features-4b}). For Gemma 3 12B IT and
Qwen3-8B we instead run an English-only medical-vs-non-medical
contrastive at each analyzed layer, using the $60$ NF prompts as
the medical set and $30$ patient-voiced non-medical prompts as
control (Appendix~\ref{app:contrastive-prompts}). Following the
mean-conditional-activation contrast~\cite{bricken2023monosemanticity,templeton2024scaling}, features
are scored by
$\mathbb{E}_\text{med}[a_f(p)] - \mathbb{E}_\text{non}[a_f(p)]$,
where $a_f(p)$ is defined in Eq.~\ref{eq:maxpool} and
$\mathbb{E}[\cdot]$
is the empirical mean of feature $f$'s per-prompt peak activation
over the medical/non-medical prompt set. Features are then
filtered for selectivity ($\geq 70\%$ medical and $\leq 10\%$
non-medical firing above threshold $1.0$) and the top-$3$ are
retained.
A $K$-sweep confirms stability across $K\in\{3,5,10,20\}$
(Appendix~\ref{app:phase1b_sensitivity}); within-corpus feature
identification at 12B and Qwen is scoped in
Section~\ref{sec:limitations}.

\section{Results}

\subsection{Behavioral differences}
\label{sec:phase0_5_cells}

We first test whether an output-format difference retains its
direction across input styles. We measure accuracy on all 60 cases
(Section~\ref{sec:dataset}; Table~\ref{tab:phase0_5_cells}).
NF and SF responses are scored by two LLM
judges %(\texttt{gpt-5.2-thinking-high} and \texttt{claude-sonnet-4.6})
under the four-letter adjudication prompt of
Section~\ref{sec:scoring}.

\begin{table}[ht]
\centering
\footnotesize
\setlength{\tabcolsep}{2.5pt}
\begin{tabular}{lrrr}
\toprule
Condition & 4B & 12B & Qwen \\
\midrule
SL (structured, MCQ)                    & 58.3\% & 81.7\% & 75.0\% \\
SF (structured, free)                   & 63.3\% & 73.3\% & 70.0\% \\
NL (natural, MCQ)                       & 55.0\% & 81.7\% & 75.0\% \\
NF (natural, free)                      & 71.7\% & 71.7\% & 68.3\% \\
\midrule
Natural input: NF$-$NL                & $+16.7$ & $-10.0$ & $-6.7$ \\
SF$-$SL (descriptive)                 & $+5.0$  & $-8.3$  & $-5.0$ \\
\bottomrule
\end{tabular}
\caption{Condition-level accuracy on the 60 cases. The SF$-$SL row
is a descriptive direction-consistency check; differences are
percentage points.}
\label{tab:phase0_5_cells}
\end{table}

\paragraph{Cross-input robustness.}
Free-text is higher at 4B ($+16.7$/$+5.0$pp for
natural/structured input) and lower at 12B ($-10.0$/$-8.3$pp) and
Qwen ($-6.7$/$-5.0$pp). Paired McNemar tests on NF$-$NL give
$p{=}0.031$ at both Gemma scales and $p{=}0.45$ at Qwen. The
descriptive SF$-$SL check has the same sign at every model but is
not significant ($p{=}0.629/0.180/0.581$ at 4B/12B/Qwen). We
therefore treat Qwen and SF$-$SL as directional consistency, not
confirmation. Per-judge results are in
Appendix~\ref{app:deferred_adjudication}.

\paragraph{NL-NF performance details.}
Under the both-judges rule, NF exceeds NL by $17$pp at 4B,
replicating the constrained-output penalty
of~\cite{frailenavarro2026triage}; the 5-way refinement finds no
unanimous deferrals. NF is instead $10$pp below NL at 12B and
$6.7$pp below at Qwen.
The 4B free-text gain is concentrated where adjacent dispositions
compete: gold $C$ (NL $10\%\to$ NF $80\%$, $+70$pp) and gold
%$C/D$ (NL $75\%\to$ NF $96\%$, $+21$pp), spanning the
$C/D$ ($+21$pp), spanning the
Intermediate and High acuity tiers. The 12B pattern is 
inverse, with NF underperforming most on gold $D$ ($-25$pp) and
gold $C$ ($-20$pp) (Table~\ref{tab:acuity-by-gold}). At Qwen, NF
is weakest on the low-acuity gold $A$ ($-24$pp; Table~\ref{tab:acuity-qwen}), consistent with the 12B
patterns. Per-acuity
breakdowns for 4B and 12B in Appendix~\ref{app:acuity-stratification}.

\paragraph{Triage error direction.}
Multiple-choice errors skew toward under-triage at 4B (NL
$20$ under/$7$ over) and Qwen ($12/3$), but toward
over-triage at 12B (SL $3/8$; NL $4/7$). NF errors are
more balanced ($5/8$, $8/9$, and $8/6$ at 4B/12B/Qwen). Full
correct/under/over/no-commit counts appear in
Appendix~\ref{app:triage-direction}.

\subsection{What accounts for the accuracy gaps?}
\label{sec:deferred_class}

We ask whether cases that differ between formats mainly reflect shifts
between nearby triage levels or free-text responses that do not commit
to a single level. Table~\ref{tab:stratum_decomp} summarises this
decomposition; per-case results are in
Appendix~\ref{app:gap_decomposition}.

\begin{table}[ht]
\centering
\footnotesize
\setlength{\tabcolsep}{4pt}
\begin{tabular}{lccccc}
\toprule
Model & NF\_OR & NL\_OR & DEF & in gap & Adj. \\
\midrule
4B   & 14 & 1 & 0  & 0    & 14/14 + 1/1 \\
12B  & 0  & 6 & 4  & 0    & 5/6 \\
Qwen & 6  & 8 & 2  & 0--2 & 1/6 + 8/8 \\
\bottomrule
\end{tabular}
\caption{Decomposition of the NL-NF accuracy gap.
NF\_OR / NL\_OR $=$ NF-only-right / NL-only-right counts;
DEF $=$ unanimous \textsc{deferred} cases under 5-way scoring;
\emph{in gap} $=$ contribution to the 4-way-scored gap (deferrals
flattening to gold-compatible letters count as correct, $0$;
Qwen's two deferrals are judges-disagree, $0$--$2$);
Adj.\ $=$ fraction of gap-driving cases that are
single-acuity-step miscalibrations.}
\label{tab:stratum_decomp}
\end{table}

\paragraph{Adjacent miscalibration accounts for the gap.}
At Gemma 4B, the 14 gap-driving cases share a single pattern: NL
picks B, gold is C, and NF picks C in $13/14$. 
%(the 14th case is NL${=}$A, gold${=}$B, NF picks B). 
In the multiple-choice mode, 4B
systematically commits to B on cases where the gold disposition is 
C, while the same model under free-text reaches the correct answer.
Exhaustive option-order shuffles (Section~\ref{sec:option-order},
Appendix~\ref{app:option_shuffle}) rule out an
alternative explanation that 4B is positionally biased toward
$B$: across all models the picked letter is at or
below chance under shuffles (case-clustered $95\%$ CI excludes
$25\%$ at 4B and 12B; contains it at Qwen), while picked
\emph{content} stays far above chance ($64.\%$, $80.3\%$, $82.6\%$
at 4B, 12B, Qwen). At 4B specifically, randomizing the
letter-to-content assignment raises NL's accuracy from
$55.0\%$ to $69.8\%$, statistically
indistinguishable from NF's $71.7\%$. Thus, 4B's gap is associated
with the original answer-key layout interacting with a content
preference, not simple positional bias: the shuffle raises 4B
accuracy to NF level while leaving 12B and Qwen mostly unchanged.
At 12B and Qwen, the canonical mapping is approximately neutral
($-2.8$ and $-2.2$pp under shuffle), but a $4.6$--$7.1$pp
difference remains after shuffling, so answer-key mapping does not
explain the full NL--NF difference.

\paragraph{Deferral does not account for the gap.}
Under the 5-way label space of
Section~\ref{sec:deferral-adjudication}, NF gives a 
conditional answer instead of committing to a letter in
$4$ cases at 12B, $2$ at Qwen, and none at 4B.
When these are re-scored under the original
4-letter space, all 12B deferrals flatten to a
gold-compatible letter and count as correct, not contributing 
to the 12B gap; the Qwen deferrals receive split
judgments from the two LLM judges (judges-disagree) and
contribute partially. A clinician judged $3$ of the 12B
deferrals clinically appropriate
(Appendix~\ref{app:clinician}). The structured-input SF
condition shows the inverse profile ($4$ unanimous deferrals
at 4B, none at Qwen): conditional responses are more common under
structured input in the 4B model, a pattern worth following up.
Answer deferral is therefore a label-space concern: the
4-letter benchmark cannot express conditional triage boundaries,
but it is not the driver of the observed accuracy inversion.

\subsection{Medical-feature responses across formats}
\label{sec:phase1b_strata}

We test the hypothesis that vignette-level medical SAE features
are more format-invariant than random features
(Section~\ref{sec:medical-feature-id}) under
both summary metrics defined in
Section~\ref{sec:mech-invariance}. We report the late layer per model (4B L29, 12B L31, Qwen L31; rationale and per-layer behavior in Section~\ref{sec:mech-invariance} and Appendix~\ref{app:full-tables}).

\begin{table}[ht]
\centering
\footnotesize
\setlength{\tabcolsep}{4pt}
\begin{tabular}{llrcc}
\toprule
Model / Layer & Stratum & $n$ & $\Delta$sMAPE & $\Delta$cos \\
\midrule
\multirow{3}{*}{4B L29}
  & both-right     & 29 & $-0.275^{\ast}$ & $+0.054^{\ast}$ \\
  & both-wrong     & 12 & $-0.336^{\ast}$ & $+0.060^{\ast}$ \\
  & NF\_OR         & 14 & $-0.338^{\ast}$ & $+0.049^{\ast}$ \\
\midrule
\multirow{3}{*}{12B L31}
  & both-right     & 43 & $-0.222^{\ast}$ & $+0.093^{\ast}$ \\
  & both-wrong     & 11 & $-0.232^{\ast}$ & $+0.094^{\ast}$ \\
  & NL\_OR         & 6  & $-0.262^{\ast}$ & $+0.093^{\ast}$ \\
\midrule
Qwen L31 & all     & 60 & $-0.070^{\ast}$ & $+0.004^{\ast}$ \\
\bottomrule
\end{tabular}
\caption{Late-layer per-stratum medical-random feature
difference. $^{\ast}$ marks cells significant at $p<0.05$
(bootstrap $95\%$ CI excludes zero, $2{,}000$ resamples). At 12B the NF-only-right cell is empty;
the 6-case NL-only-right stratum substitutes (the gap-driver at
this scale; Section~\ref{sec:deferred_class}). Per-layer /
per-stratum tables in Appendix~\ref{app:full-tables}; sanity
checks in Appendices~\ref{app:token_masks}
and~\ref{app:resample}.}
\label{tab:phase1b_headline}
\end{table}

\paragraph{Medical features.}
$\Delta$sMAPE${<}0$ and $\Delta$cos${>}0$ both indicate medical
features are more format-invariant than the magnitude-matched
random control. At every cell of 4B L29 and 12B L31 the difference
is significant on both metrics (bootstrap $95\%$ CIs non-zero). Qwen L31 reproduces the direction at smaller effect
size, consistent with Qwen-Scope's higher reconstruction error
(Section~\ref{sec:models-saes}).

A natural concern is that NL and NF
share the same clinical vignette verbatim, and tokens inside that
shared block can only attend leftward, so their
hidden states are identical across formats. The invariance we
report is different. Under max-pool, the medical
features' peak activations actually \emph{lie inside} the shared
clinical vignette in $98.3$--$100\%$ of (case, feature) combinations
across all three models, whereas the magnitude-matched random
features peak on the NL-only scaffold tokens. The medical features
are tracking clinical content; the random pool is not.

To rule out the mechanical reading we conduct additional checks.
\emph{First}, a vignette-only token mask restricts both pools to 
tokens inside the shared clinical block. sMAPE drops to
$0.002$-$0.006$ for medical {and} random features alike
(Appendix~\ref{app:token_masks}), confirming that the observed format
contrast occurs outside the vignette and the
medical-random gap would not arise from prefix-attention
alone. \emph{Second}, $1{,}000$ magnitude-matched random resamples 
preserve the medical-random gap at $p{<}0.001$ at every model
(Appendix~\ref{app:resample}). \emph{Third}, restricting the 
random pool to features that fire on $\geq25\%$ of the
NL${\cup}$NF prompts, so the comparison is against
content-firing features, %not arbitrary low-activation noise) 
leaves every stratum's $95\%$ CI non-zero on both metrics, with
modest magnitude changes relative to the unrestricted control
%(at 4B L29 NF-only-right $-0.338\to-0.272$; 
(per-stratum
results in Appendix~\ref{app:full-tables}, where the Gemma 12B
shallow/mid sign inversion at L12/L24 is shown). Taken together,
the identified medical features respond similarly across formats on
the clinical narrative; this does not establish that the model
represents the correct triage disposition.

\paragraph{Direction of the residual difference.}
\label{sec:direction_analysis}
Where does the format direction align in the SAE? We project
the case-averaged NL--NF residual onto SAE
encoder columns under the max-pool aggregation of
Section~\ref{sec:direction-test} and rank features by absolute
alignment. Across all three models, the medical features land at mid-rank percentiles (median $47\%$, IQR $33$--$68\%$, with $11/12$ Gemma
layer-feature combinations outside the top decile). The format
direction therefore aligns most strongly with \emph{non-medical}
features. The pattern is consistent across the three aggregations
(full mean-pool, length-controlled mean-pool, and max-pool;
Section~\ref{sec:direction-test}), and top-token inspection of the
most-aligned 4B L29 features shows them firing on NL-only
answer-key scaffold tokens. Full
ranks and the per-aggregation comparison are in
Appendices~\ref{app:full-tables} and~\ref{app:format_features}.

%\paragraph{Input-style robustness.}
\label{sec:phase1b_robustness}
Repeating the medical-random invariance test on the
structured-input SL-SF pair at the same
layers yields paired $\Delta$sMAPE values of $-0.081$ at 4B L29,
$-0.062$ at 12B L31, and $-0.150$ at Qwen L31, all with $95\%$ CI
strictly non-zero. The same directional association therefore
appears under both input styles. Full table in
Appendix~\ref{app:input-style-robustness}.

\subsection{Acuity information across positions}
\label{sec:acuity-probes}

Stable medical-domain features do not necessarily mean that the model
retains information about how urgent the case is. We therefore test
whether emergency status remains decodable across formats. At the
vignette, the emergency-tier probe reaches
ROC-AUC $0.954/0.962$ (NL/NF) at 4B, $0.993/0.994$ at 12B, and
$0.999/1.000$ at Qwen (all permutation $p{=}0.001$). Within each
model, the format difference is at most $0.009$ and every paired
$95\%$ CI spans zero; four-class most-urgent-tier macro-OVR AUC is
$0.976$--$0.993$.

\begin{figure*}[t]
  \centering
  \includegraphics[width=\textwidth]{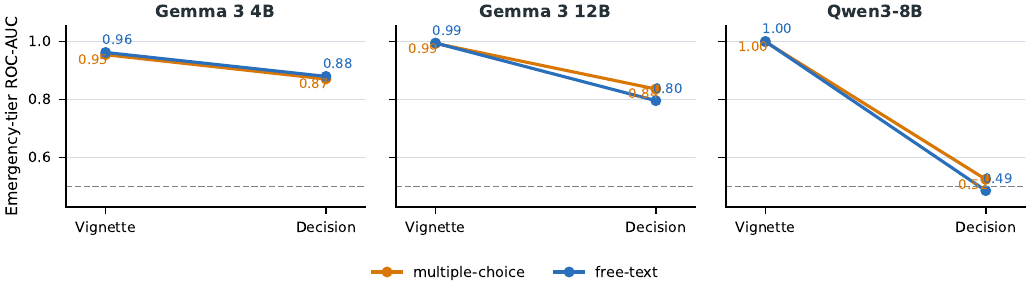}
  \caption{Emergency-tier information is similarly linearly decodable
  across formats at the vignette, then attenuates at the decision
  token. Lines connect vignette and decision-token ROC-AUC for each
  output format; dashed line is chance. The paired drop is
  significant at 12B and Qwen (case-bootstrap $95\%$ CI excludes
  zero) but not at 4B.}
  \label{fig:acuity-probes}
\end{figure*}

At the decision token, AUC is $0.871/0.879$ at 4B,
$0.836/0.796$ at 12B, and $0.525/0.485$ at Qwen. The
vignette-to-decision drop is significant at 12B
($0.157/0.199$) and Qwen ($0.474/0.515$), and directional but not
significant at 4B ($0.084/0.083$). Thus 4B retains substantial
urgency signal at the decision token, whereas Qwen falls to chance. These probes
establish linear decodability at the analyzed layer, not causal use
of the signal. The probe script, permutation outputs, and paired
bootstrap results are included in the released repository.

\subsection{What changes at answer selection?}
\label{sec:decision_token}

We test whether the format-associated representation difference
concentrates where the answer is selected. We turn to the NL
pre-generation
%The question this subsection answers is what the residual stream contains at the single 
token, where the multiple-choice letter is decided.
%whose hidden state the model head reads to emit the first letter. 
Our primary tool here is the Natural Language Autoencoder
(NLA)~\citet{frasertaliente2026nla}, an unsupervised method
\emph{independent of SAE} that produces
natural-language descriptions of residual-stream activations. We
capture L32 activations of Gemma 3 12B IT at seven token positions
under NL and NF (Section~\ref{sec:methods_decision_token},
Appendix~\ref{app:nla-details}) and the LLM judges classify
each description's \textsc{medical} and \textsc{scaffold} grade, 
with three classes each: \textsc{primary} -- description's main 
subject, \textsc{partial} -- secondary mention, or \textsc{no} -- 
or absent. The judges agree $96.7\%$ and $96.2\%$ of time 
on the respective axes (Table~\ref{tab:phase8_nla}).

\begin{table}[ht]
\centering
\scriptsize
\setlength{\tabcolsep}{2pt}
\begin{tabular}{lcccccc}
\toprule
\multirow{2}{*}{Position} & \multicolumn{3}{c}{\textsc{medical} ($n{=}60$)} & \multicolumn{3}{c}{\textsc{scaffold} ($n{=}60$)} \\
\cmidrule(lr){2-4}\cmidrule(lr){5-7}
 & \textsc{prim} & \textsc{part} & \textsc{no} & \textsc{prim} & \textsc{part} & \textsc{no} \\
\midrule
\rowcolor[gray]{0.92} NF\_content   & \textbf{60} & 0 & 0           & 0           & 8 & \textbf{52} \\
NL\_content                         & 60          & 0 & 0           & 0           & 7 & 53          \\
\rowcolor[gray]{0.92} NL\_decision   & 0          & \textbf{60} & 0 & \textbf{60} & 0 & 0           \\
NL\_letter\_A                       & 0           & 60 & 0           & 60          & 0 & 0           \\
NL\_letter\_B                       & 0           & 8  & \textbf{52} & 60          & 0 & 0           \\
NL\_letter\_C                       & 0           & 57 & 3           & 60          & 0 & 0           \\
NL\_letter\_D                       & 0           & 60 & 0           & 60          & 0 & 0           \\
\bottomrule
\end{tabular}
\caption{LLM-judge classifications of NLA descriptions across $7$
token positions, with headline cells shaded.
NF\_content/NL\_content: clinical content is the \textsc{med-primary}
frame under both formats. NL\_decision: the content frame is
demoted to \textsc{med-partial} and a \textsc{sca-primary}
multiple-choice-scaffold encoding takes over.}
\label{tab:phase8_nla}
\end{table}

\paragraph{NLA descriptions shift at the decision token.}
At {NF\_content} and {NL\_content}, the NLA describes clinical
content as the primary subject in all cases. At {NL\_decision},
after the multiple-choice scaffold, all descriptions instead classify
the scaffold as primary and medical content as secondary
(Table~\ref{tab:phase8_nla}). The
multiple-choice scaffold does not strip clinical content, but
\emph{demotes} that content from the representation's main
subject to a secondary mention while a multiple-choice-scaffold
concept becomes the primary subject.
Representative NLA outputs are in Appendix~\ref{app:nla-details}.

\paragraph{SAE-pool corroboration.}
Two diagnostics on the SAE feature pool corroborate the NLA reading.
\emph{First},
the three contrastively-identified medical features used
in Section~\ref{sec:phase1b_strata} have no activation
at the NL pre-generation token in $60$ cases at every model,
and contribute $0\%$ of the absolute linear effect on
the A/B/C/D logits via the per-feature unembedding projection
(Appendix~\ref{app:logit_attribution}). \emph{Second}, 
top-$20$ active features at the NL and NF decision tokens are 
essentially disjoint at Gemma %(Jaccard ${\approx}\,0$)
and share $32\%$ at Qwen. Across all models,
$87$-$95\%$ of features unique to NL's top-$20$ peak on scaffold
tokens outside the shared clinical vignette and no
identified medical features appear in either top-$20$ at any
model (Appendix~\ref{app:decision_token_features}). These 
results concur with NLA: medical-content detectors fire on the
clinical narrative during reading, then go silent before the letter
is selected. In the exact all-active-feature projection,
scaffold-peaking features account for $93.2\%$ (4B) and $91.9\%$
(12B) of unsigned attribution and $96.1\%/96.0\%$ of positive
predicted-letter attribution; the identified medical features
contribute zero. These are linear projections, not causal effects
(Appendices~\ref{app:logit_attribution}
and~\ref{app:decision_token_features}).

\subsection{Source-format states predict which cases change}
\label{sec:probe_results}

We test whether the source-format hidden state predicts which cases
flip under a format change, as expected if decision-stage
representations covary with the behavioral gap. We test this on
Gemma 3 4B IT.
\begin{figure}[h]
    \centering
    \includegraphics[width=\columnwidth]{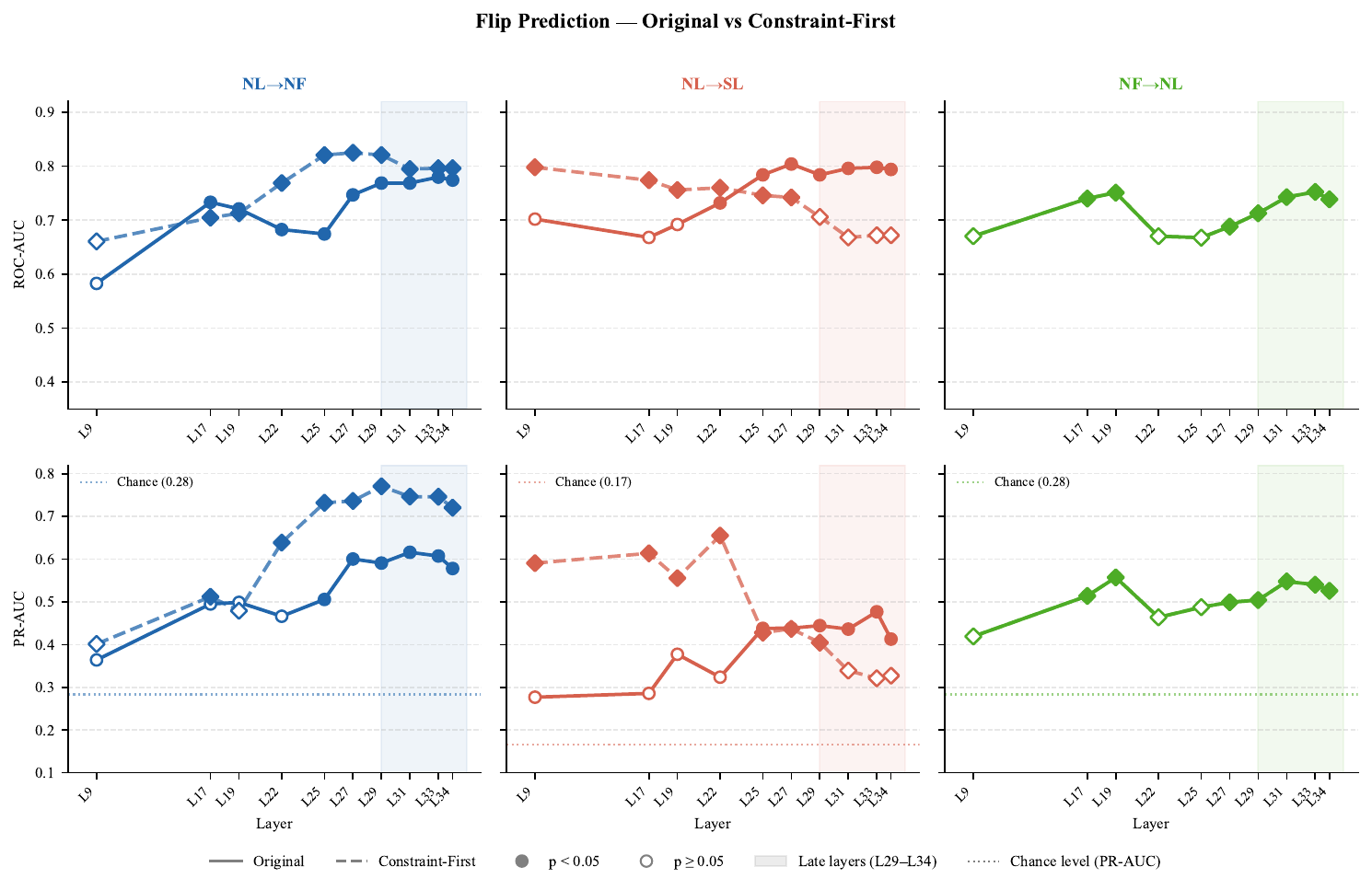}
    \caption{Linear probe AUC and PR-AUC performance with p-values for $\text{NL}\rightarrow \text{NF}$, $\text{NL}\rightarrow \text{SL}$, $\text{NF}\rightarrow \text{NL}.$}
    \label{fig:flip_prediction}
\end{figure}

Predictive signal concentrates in the deep layers of the network (L29-L34). For $\text{NL}\rightarrow\text{NF}$, we obtain AUC=$0.769-0.780$, $p=0.004$ and PR-AUC=$0.591-0.616$ vs baseline $0.283$; for $\text{NL}\rightarrow\text{SL}$ -- AUC=$0.784-0.804$, $p=0.010$ and PR-AUC=$0.413-0.477$ vs baseline $0.167$; and for $\text{NF}\rightarrow\text{NL}$ -- AUC=$0.713-0.752$, $p=0.009-0.023$ (see Figure \ref{fig:flip_prediction}). Among the four source prompt formats, NL-sourced last-token embeddings produce the most consistent predictive signal across target directions at late layers. All the results are in Appendix \ref{app:supplementary layer-by-layer probing results}. 

%\paragraph{Constraint-First ablation.}
Moving the formatting instruction to the beginning of the prompt  in constraint-first ablation selectively modifies the late-layer predictive signal for the two NL-sourced transitions. For $\text{NL\_{CF}}\rightarrow\text{NF}$, AUC at the same levels increases by $0.016-0.052$ ($p=0.001-0.003$), while for $\text{NL\_{CF}}\rightarrow\text{SL}$ it degrades by $0.078-0.128$ and loses significance. NF-sourced directions were identical across both runs since NF does not have output constraints. 

\section{Conclusion}
\label{sec:conclusion}

Across three instruction-tuned LLMs from two families, the strongest
correlates of the multiple-choice format effect appear near answer
selection. Medical features and urgency-related information remain
similar across formats during vignette processing. At the decision
token, medical features are inactive and urgency weakens. NLA and
top-feature analysis associate the state with the multiple-choice
scaffold; exact linear projection assigns over $91\%$ of unsigned
attribution to scaffold-peaking features and zero to the identified
medical features. The effect varies by model: option shuffling shows
that the 4B result depends on the original answer-key mapping, and
recommendations that differ between formats are usually one triage
level apart rather than deferrals.
These findings support an output-mapping account, while leaving open
which internal mechanisms causally produce the behavioral differences.

\paragraph{Availability.}
The full corpus, code, judge prompts, raw model and judge outputs,
and analysis scripts are available at
\url{https://github.com/dafraile/SAE_mad}.

\section{Limitations}
\label{sec:limitations}

\paragraph{Coverage and sample size.} We study 60 clinical-triage
cases per model across three instruction-tuned LLMs from two families,
with greedy decoding only. Small
behavioral strata and the one-layer Qwen analysis limit
generalization beyond Gemma and Qwen. The Qwen behavioral gap is
not significant at $n{=}60$ (McNemar $p{=}0.45$); the Qwen
mechanistic findings are nonetheless directionally consistent at
smaller effect size, consistent with Qwen-Scope's higher
reconstruction error on the residual stream.

\paragraph{Judging and clinical calibration.} NF scoring relies on
two LLM judges under an A--D label space that flattens some tiered
\textsc{deferred} responses. The clinician subset is enriched and
small ($n{=}16$, Appendix~\ref{app:clinician}); it calibrates label
face validity, not clinical deferral prevalence or deployment
safety.

\paragraph{Mechanistic-method scope.} We use max/mean pooling and
encoder-direction projection rather than richer aggregations or a
full multi-feature intervention sweep. Qwen-Scope's
${\sim}34\%$ L31 reconstruction error raises the Qwen noise floor.
Feature identification at 12B and Qwen reuses the 60 NF prompts
with a 30-prompt non-medical contrastive; a held-out contrastive
split would tighten inference. The random baseline is
magnitude-matched and resampled (Appendix~\ref{app:resample}) to
avoid zero-firing denominator artifacts.

\paragraph{Claim scope.} The contrastively selected SAE features
are medical-vs-non-medical detectors; acuity is tested separately
with supervised linear probes. High leave-one-out AUC establishes
linear decodability at one analyzed layer on these 60 cases, not
that the model uses the signal or that it will generalize to a new
cohort. Because NL and NF share the same vignette prefix, some
similarity is guaranteed by causal masking; token-mask and peak-location
checks separate this trivial overlap from the reported feature-level
results (Appendix~\ref{app:token_masks}). NLA evidence is limited to Gemma
3 12B IT, the released checkpoint available for this analysis
~\cite{frasertaliente2026nla}. Causal interventions on the
format-direction features at 4B L29 (discrete SAE-feature ablation
and continuous ActAdd steering;
Appendix~\ref{app:intervention-details}) change only $0$--$2$ of
$60$ NL predictions, consistent with the multi-intervention nulls
of~\cite{basu2026interpretability}; we therefore treat the SAE
features as \emph{readout probes} of medical-domain content
rather than as causal handles or clinical monitors. Activation
patching between matched decision-token states is the planned next
causal test.

\paragraph{Decision-token feature sets differ substantially across
formats.} At the NL pre-generation token, the
top-$20$ most active features are nearly disjoint between NL and
NF at both Gemma scales (Jaccard $\approx 0$; $0.32$ at Qwen), and
\emph{none} of the contrastively-identified medical features
appear in either set
(Appendix~\ref{app:decision_token_features}). Thus the most-active
decision-token feature sets differ strongly at Gemma and partially at
Qwen. We do not yet know what the scaffold-peaking features encode
beyond their peak locations, nor
whether they reflect learned answer-key templates from
pretraining/instruction tuning or something more specific to the
clinical-triage register; characterising them is left to future work.

\paragraph{Clinical claim scope.} Lower forced-letter under-triage
in some comparisons is not a safety endorsement of NF. Deferrals,
hedging, and emergency-tier misses persist; our claim concerns the
adequacy of forced-letter evaluation as a competence readout, not
the deployment-readiness of free-text outputs. Late-layer SAE
features are a \emph{candidate readout} of medical-domain
representation, not a validated clinical monitor: deployment would
require held-out calibration, robustness checks, correlation with
under-triage risk, and prospective validation. The intervention
nulls here and in~\cite{basu2026interpretability} also argue
against single-layer SAE control as a correction method on this
task.

\section{Use of AI assistants}

We used AI assistants---Claude Code and Claude Opus 4.7
(Anthropic), and Codex (OpenAI)---at three stages of this work:
(i) implementing analysis code (SAE
encoding, bootstrap and case-clustered CI routines, option-order
shuffles, decision-token logit attribution, the NLA
judge-classification pipeline, and the linear-probe training);
(ii) executing experiments under human supervision (model
loading, hooked forward passes, activation extraction, judge-call
orchestration), with authors verifying outputs against unit tests
and independent reproductions where applicable; and (iii) paper
preparation (LaTeX editing, prose tightening on the basis of
co-author and reviewer comments, bibliography reconciliation).
Research design, hypotheses, method and model selection, and
interpretation of results were determined by the human authors.
AI-assisted prose and LaTeX changes were reviewed against the
underlying results and approved by the authors, who take full
responsibility for all numerical results, code, and claims.

\bibliography{references}

\begin{thebibliography}{27}
\providecommand{\natexlab}[1]{#1}

\bibitem[{Basu et~al.(2026)Basu, Patel, Sheth, Muralidharan, Elamaran, Kinra, Morgan, and Batniji}]{basu2026interpretability}
Sanjay Basu, Sadiq~Y. Patel, Parth Sheth, Bhairavi Muralidharan, Namrata Elamaran, Aakriti Kinra, John Morgan, and Rajaie Batniji. 2026.
\newblock \href {https://arxiv.org/abs/2603.18353} {Interpretability without actionability: mechanistic methods cannot correct language model errors despite near-perfect internal representations}.
\newblock \emph{arXiv preprint arXiv:2603.18353}.
\newblock Submitted 2026-03-18 (verified 2026-05-11).

\bibitem[{Bricken et~al.(2023)Bricken, Templeton, Batson, Chen, Jermyn, Conerly, Turner, Anil, Denison, Askell, Lasenby, Wu, Kravec, Schiefer, Maxwell, Joseph, Hatfield-Dodds, Tamkin, Nguyen, McLean, Burke, Hume, Carter, Henighan, and Olah}]{bricken2023monosemanticity}
Trenton Bricken, Adly Templeton, Joshua Batson, Brian Chen, Adam Jermyn, Tom Conerly, Nick Turner, Cem Anil, Carson Denison, Amanda Askell, Robert Lasenby, Yifan Wu, Shauna Kravec, Nicholas Schiefer, Tim Maxwell, Nicholas Joseph, Zac Hatfield-Dodds, Alex Tamkin, Karina Nguyen, and 6 others. 2023.
\newblock Towards monosemanticity: Decomposing language models with dictionary learning.
\newblock \url{https://transformer-circuits.pub/2023/monosemantic-features}.

\bibitem[{Burns et~al.(2023)Burns, Ye, Klein, and Steinhardt}]{burns2023discovering}
Collin Burns, Haotian Ye, Dan Klein, and Jacob Steinhardt. 2023.
\newblock \href {https://arxiv.org/abs/2212.03827} {Discovering latent knowledge in language models without supervision}.
\newblock \emph{International Conference on Learning Representations}.

\bibitem[{Cunningham et~al.(2023)Cunningham, Ewart, Riggs, Huben, and Sharkey}]{cunningham2023sparse}
Hoagy Cunningham, Aidan Ewart, Logan Riggs, Robert Huben, and Lee Sharkey. 2023.
\newblock \href {https://arxiv.org/abs/2309.08600} {Sparse autoencoders find highly interpretable features in language models}.
\newblock \emph{arXiv preprint arXiv:2309.08600}.

\bibitem[{Fraile~Navarro(2026)}]{frailenavarro2026saemad}
David Fraile~Navarro. 2026.
\newblock Sae-guided cross-lingual knowledge transfer in llms: A null result with preserved findings.
\newblock \url{https://github.com/dafraile/SAE_mad}.

\bibitem[{Fraile~Navarro et~al.(2026)Fraile~Navarro, Magrabi, and Coiera}]{frailenavarro2026triage}
David Fraile~Navarro, Farah Magrabi, and Enrico Coiera. 2026.
\newblock \href {https://arxiv.org/abs/2603.11413} {Evaluation format, not model capability, drives triage failure in the assessment of consumer health {AI}}.
\newblock \emph{arXiv preprint arXiv:2603.11413}.
\newblock Submitted 2026-03-12, v3 revised 2026-03-26 (verified 2026-05-11).

\bibitem[{Fraser-Taliente et~al.(2026)Fraser-Taliente, Kantamneni, Ong, Mossing, Lu, Bogdan, Ameisen, Chen, Kishylau, Pearce, Tarng, Wu, Wu, Zhang, Ziegler, Hubinger, Batson, Lindsey, Zimmerman, and Marks}]{frasertaliente2026nla}
Kit Fraser-Taliente, Subhash Kantamneni, Euan Ong, Dan Mossing, Christina Lu, Paul~C. Bogdan, Emmanuel Ameisen, James Chen, Dzmitry Kishylau, Adam Pearce, Julius Tarng, Alex Wu, Jeff Wu, Yang Zhang, Daniel~M. Ziegler, Evan Hubinger, Joshua Batson, Jack Lindsey, Samuel Zimmerman, and Samuel Marks. 2026.
\newblock Natural language autoencoders produce unsupervised explanations of {LLM} activations.
\newblock \url{https://transformer-circuits.pub/2026/nla/index.html}.
\newblock Anthropic; published 2026-05-07. First three authors contributed equally (listed alphabetically among them).

\bibitem[{Gao et~al.(2024)Gao, Dupr{\'e}~la Tour, Tillman, Goh, Troll, Radford, Sutskever, Leike, and Wu}]{gao2024scaling}
Leo Gao, Tom Dupr{\'e}~la Tour, Henk Tillman, Gabriel Goh, Rajan Troll, Alec Radford, Ilya Sutskever, Jan Leike, and Jeffrey Wu. 2024.
\newblock \href {https://arxiv.org/abs/2406.04093} {Scaling and evaluating sparse autoencoders}.
\newblock \emph{arXiv preprint arXiv:2406.04093}.

\bibitem[{Griot et~al.(2025)Griot, Vanderdonckt, Yuksel, and Hemptinne}]{griot2025pattern}
Maxime Griot, Jean Vanderdonckt, Demet Yuksel, and Coralie Hemptinne. 2025.
\newblock \href {https://doi.org/10.18653/v1/2025.acl-long.266} {Pattern recognition or medical knowledge? the problem with multiple-choice questions in medicine}.
\newblock In \emph{Proceedings of the 63rd Annual Meeting of the Association for Computational Linguistics (Volume 1: Long Papers)}, pages 5321--5341, Vienna, Austria. Association for Computational Linguistics.

\bibitem[{Kadavath et~al.(2022)Kadavath, Conerly, Askell, Henighan, Drain, Perez, Schiefer, Hatfield-Dodds, DasSarma, Tran-Johnson, Johnston, El-Showk, Jones, Elhage, Hume, Chen, Bai, Bowman, Fort, Ganguli, Hernandez, Jacobson, Kernion, Kravec, Lovitt, Ndousse, Olsson, Ringer, Amodei, Brown, Clark, Joseph, Mann, McCandlish, Olah, and Kaplan}]{kadavath2022language}
Saurav Kadavath, Tom Conerly, Amanda Askell, Tom Henighan, Dawn Drain, Ethan Perez, Nicholas Schiefer, Zac Hatfield-Dodds, Nova DasSarma, Eli Tran-Johnson, Scott Johnston, Sheer El-Showk, Andy Jones, Nelson Elhage, Tristan Hume, Anna Chen, Yuntao Bai, Sam Bowman, Stanislav Fort, and 17 others. 2022.
\newblock \href {https://arxiv.org/abs/2207.05221} {Language models (mostly) know what they know}.
\newblock \emph{arXiv preprint arXiv:2207.05221}.

\bibitem[{Lieberum et~al.(2024)Lieberum, Rajamanoharan, Conmy, Smith, Sonnerat, Varma, Kram{\'a}r, Dragan, Shah, and Nanda}]{lieberum2024gemma}
Tom Lieberum, Senthooran Rajamanoharan, Arthur Conmy, Lewis Smith, Nicolas Sonnerat, Vikrant Varma, J{\'a}nos Kram{\'a}r, Anca Dragan, Rohin Shah, and Neel Nanda. 2024.
\newblock \href {https://aclanthology.org/2024.blackboxnlp-1.19/} {Gemma scope: Open sparse autoencoders everywhere all at once on gemma 2}.
\newblock In \emph{Proceedings of the 7th BlackboxNLP Workshop: Analyzing and Interpreting Neural Networks for NLP}, pages 217--232.

\bibitem[{Makelov et~al.(2024)Makelov, Lange, and Nanda}]{makelov2024principled}
Aleksandar Makelov, George Lange, and Neel Nanda. 2024.
\newblock \href {https://arxiv.org/abs/2405.08366} {Towards principled evaluations of sparse autoencoders for interpretability and control}.
\newblock \emph{arXiv preprint arXiv:2405.08366}.

\bibitem[{Makhzani and Frey(2013)}]{makhzani2013ksparse}
Alireza Makhzani and Brendan Frey. 2013.
\newblock \href {https://arxiv.org/abs/1312.5663} {k-sparse autoencoders}.
\newblock \emph{arXiv preprint arXiv:1312.5663}.

\bibitem[{Makridakis(1993)}]{makridakis1993accuracy}
Spyros Makridakis. 1993.
\newblock \href {https://doi.org/10.1016/0169-2070(93)90079-3} {Accuracy measures: theoretical and practical concerns}.
\newblock \emph{International Journal of Forecasting}, 9(4):527--529.

\bibitem[{Marks et~al.(2024)Marks, Rager, Michaud, Belinkov, Bau, and Mueller}]{marks2024sparse}
Samuel Marks, Can Rager, Eric~J. Michaud, Yonatan Belinkov, David Bau, and Aaron Mueller. 2024.
\newblock \href {https://arxiv.org/abs/2403.19647} {Sparse feature circuits: Discovering and editing interpretable causal graphs in language models}.
\newblock \emph{arXiv preprint arXiv:2403.19647}.

\bibitem[{Pezeshkpour and Hruschka(2024)}]{pezeshkpour2024large}
Pouya Pezeshkpour and Estevam Hruschka. 2024.
\newblock \href {https://arxiv.org/abs/2308.11483} {Large language models sensitivity to the order of options in multiple-choice questions}.
\newblock In \emph{Findings of the Association for Computational Linguistics: NAACL 2024}.

\bibitem[{{Qwen Team}(2026)}]{qwen2026scope}
{Qwen Team}. 2026.
\newblock Qwen-scope: Decoding intelligence, unleashing potential.
\newblock \url{https://qwen.ai/blog?id=qwen-scope}.
\newblock Accessed 2026-05-02.

\bibitem[{Rajamanoharan et~al.(2025)Rajamanoharan, Lieberum, Sonnerat, Conmy, Varma, Kram{\'a}r, and Nanda}]{rajamanoharan2024jumping}
Senthooran Rajamanoharan, Tom Lieberum, Nicolas Sonnerat, Arthur Conmy, Vikrant Varma, J{\'a}nos Kram{\'a}r, and Neel Nanda. 2025.
\newblock \href {https://arxiv.org/abs/2407.14435} {Jumping ahead: Improving reconstruction fidelity with jumprelu sparse autoencoders}.
\newblock In \emph{International Conference on Learning Representations}.

\bibitem[{Ramaswamy et~al.(2026)Ramaswamy, Tyagi, Hugo, Jiang, Jayaraman, Jangda, Te, Kaplan, Lampert, Freeman, Gavin, Tewari, Sakhuja, Naved, Charney, Omar, Gorin, Klang, and Nadkarni}]{ramaswamy2026chatgpt}
Ashwin Ramaswamy, Alvira Tyagi, Hannah Hugo, Joy Jiang, Pushkala Jayaraman, Mateen Jangda, Alexis~E. Te, Steven~A. Kaplan, Joshua Lampert, Robert Freeman, Nicholas Gavin, Ashutosh~K. Tewari, Ankit Sakhuja, Bilal Naved, Alexander~W. Charney, Mahmud Omar, Michael~A. Gorin, Eyal Klang, and Girish~N. Nadkarni. 2026.
\newblock \href {https://doi.org/10.1038/s41591-026-04297-7} {Chatgpt health performance in a structured test of triage recommendations}.
\newblock \emph{Nature Medicine}, 32(5):1671--1675.

\bibitem[{Sclar et~al.(2024)Sclar, Choi, Tsvetkov, and Suhr}]{sclar2024quantifying}
Melanie Sclar, Yejin Choi, Yulia Tsvetkov, and Alane Suhr. 2024.
\newblock \href {https://arxiv.org/abs/2310.11324} {Quantifying language models' sensitivity to spurious features in prompt design or: How i learned to start worrying about prompt formatting}.
\newblock In \emph{International Conference on Learning Representations}.

\bibitem[{Singhal et~al.(2023)Singhal, Azizi, Tu, Mahdavi, Wei, Chung, Scales, Tanwani, Cole-Lewis, Pfohl, Payne, Seneviratne, Gamble, Kelly, Babiker, Sch{\"a}rli, Chowdhery, Mansfield, Demner-Fushman, Ag{\"u}era~y Arcas, Webster, Corrado, Matias, Chou, Gottweis, Tomasev, and Liu}]{singhal2023large}
Karan Singhal, Shekoofeh Azizi, Tao Tu, S.~Sara Mahdavi, Jason Wei, Hyung~Won Chung, Nathan Scales, Ajay Tanwani, Heather Cole-Lewis, Stephen Pfohl, Perry Payne, Martin Seneviratne, Paul Gamble, Chris Kelly, Abdelrahman Babiker, Nathanael Sch{\"a}rli, Aakanksha Chowdhery, Philip Mansfield, Dina Demner-Fushman, and 8 others. 2023.
\newblock \href {https://doi.org/10.1038/s41586-023-06291-2} {Large language models encode clinical knowledge}.
\newblock \emph{Nature}, 620(7974):172--180.

\bibitem[{Singhal et~al.(2025)Singhal, Tu, Gottweis, Sayres, Wulczyn, Amin, Hou, Clark, Pfohl, Cole-Lewis et~al.}]{singhal2025expert}
Karan Singhal, Tao Tu, Juraj Gottweis, Rory Sayres, Ellery Wulczyn, Mohamed Amin, Le~Hou, Kevin Clark, Stephen~R. Pfohl, Heather Cole-Lewis, and 1 others. 2025.
\newblock \href {https://doi.org/10.1038/s41591-024-03423-7} {Toward expert-level medical question answering with large language models}.
\newblock \emph{Nature Medicine}, 31(3):943--950.

\bibitem[{Templeton et~al.(2024)Templeton, Conerly, Marcus, Lindsey, Bricken, Chen, Pearce, Citro, Ameisen, Jones, Cunningham, Turner, McDougall, MacDiarmid, Freeman, Sumers, Rees, Batson, Jermyn, Carter, and Olah}]{templeton2024scaling}
Adly Templeton, Tom Conerly, Jonathan Marcus, Jack Lindsey, Trenton Bricken, Brian Chen, Adam Pearce, Craig Citro, Emmanuel Ameisen, Andy Jones, Hoagy Cunningham, Nicholas~L. Turner, Callum McDougall, Monte MacDiarmid, C.~Daniel Freeman, Theodore~R. Sumers, Edward Rees, Joshua Batson, Adam Jermyn, and 2 others. 2024.
\newblock Scaling monosemanticity: Extracting interpretable features from claude 3 sonnet.
\newblock \url{https://transformer-circuits.pub/2024/scaling-monosemanticity/}.

\bibitem[{Turner et~al.(2023)Turner, Thiergart, Leech, Udell, Vazquez, Mini, and MacDiarmid}]{turner2023steering}
Alexander~Matt Turner, Lisa Thiergart, Gavin Leech, David Udell, Juan~J. Vazquez, Ulisse Mini, and Monte MacDiarmid. 2023.
\newblock \href {https://arxiv.org/abs/2308.10248} {Steering language models with activation engineering}.
\newblock \emph{arXiv preprint arXiv:2308.10248}.

\bibitem[{Turpin et~al.(2023)Turpin, Michael, Perez, and Bowman}]{turpin2023language}
Miles Turpin, Julian Michael, Ethan Perez, and Samuel~R. Bowman. 2023.
\newblock \href {https://arxiv.org/abs/2305.04388} {Language models don't always say what they think: Unfaithful explanations in chain-of-thought prompting}.
\newblock In \emph{Advances in Neural Information Processing Systems}.

\bibitem[{Yancey and O'Rourke(2023)}]{yancey2023emergency}
Charles~C. Yancey and Maria~C. O'Rourke. 2023.
\newblock \href {https://www.ncbi.nlm.nih.gov/books/NBK557583/} {Emergency department triage}.
\newblock In \emph{StatPearls [Internet]}. StatPearls Publishing, Treasure Island, FL.

\bibitem[{Zheng et~al.(2024)Zheng, Zhou, Meng, Zhou, and Huang}]{zheng2024large}
Chujie Zheng, Hao Zhou, Fandong Meng, Jie Zhou, and Minlie Huang. 2024.
\newblock \href {https://arxiv.org/abs/2309.03882} {Large language models are not robust multiple choice selectors}.
\newblock In \emph{International Conference on Learning Representations}.

\end{thebibliography}

\appendix

\section{Dataset and feature identification}

\subsection{Dataset label distribution}
\label{app:dataset}

The source corpus populates all four cells of the $2\times2$ input
$\times$ output factorial --- SL (structured + multiple-choice),
NL (natural + multiple-choice), SF (structured + free-text), and
NF (natural + free-text). Gold-label counts are
$\{A{:}8, B{:}8, C{:}10, D{:}4, A/B{:}2, B/C{:}4, C/D{:}24\}$. Gold
labels $\{A,B,C,D\}$ and condition codes
$\{\text{SL},\text{NL},\text{SF},\text{NF}\}$ use disjoint alphabets
to avoid ambiguity.

\subsection{Medical-feature identification at Gemma 3 4B IT}
\label{app:medical-features-4b}

The three medical features used at Gemma 3 4B IT, L29
($f_1=12570$, $f_2=893$, $f_3=12845$) were identified in our earlier
work via a 6-condition cross-lingual contrastive (3 languages
$\times$ 2 domains: \{English, Spanish, French\} $\times$ \{medical,
non-medical\}) on MMLU question stems, validated by three independent
tests: (i) top-activating token inspection on free-form medical text
not in MMLU format; (ii) feature ablation on 4B's MMLU accuracy
across all six cells, showing the medical cells drop substantially
while non-medical cells are unaffected; (iii) non-MCQ free-form
medical-text firing, confirming the features are not MCQ-format
detectors. The MMLU medical subjects used were anatomy, clinical
knowledge, college medicine, medical genetics, and professional
medicine; the non-medical control subjects were philosophy, world
religions, and global facts (\texttt{cais/mmlu} configs for English,
\texttt{openai/MMMLU} configs for Spanish and French).

For Gemma 3 12B IT and Qwen3-8B we use a simpler English-only
medical-vs-non-medical contrastive at each analyzed layer. The
asymmetry is one of research history rather than principle: the
cross-lingual validation in 4B established that the candidate
features generalize across languages and contrast groups, so for the
larger models we use the more economical English-only contrastive
described in Section~\ref{sec:medical-feature-id}.

\subsection{Non-medical contrastive prompts}
\label{app:contrastive-prompts}

The 30 hand-curated, patient-voiced non-medical prompts used as the
non-medical control set for the English-only contrastive at Gemma 3
12B IT and Qwen3-8B. The prompts match the conversational register of
the 60 NF prompts (a person describing a non-medical concern and
asking for guidance) so that the contrast isolates clinical content
rather than register.

\begin{enumerate}\small
\setlength\itemsep{0pt}
\item ``Hi, I just got a new puppy and I'm not sure how often I should be feeding her. She's 8 weeks old, a Labrador. Any guidance on a feeding schedule?''
\item ``I'm a 30-year-old who's never done any real cooking. I want to start learning to make basic dinners. Where should I begin?''
\item ``Hey, I'm planning a trip to Tokyo next month for two weeks. I've never been to Japan. What should I prioritize seeing, and is two weeks enough?''
\item ``Hi, I just moved into my first apartment and I have no idea how to handle laundry properly. Can you walk me through the basics?''
\item ``I bought a sourdough starter last week. How often should I feed it, and can I keep it in the fridge between bakes?''
\item ``Hi, I'm 26 and want to start saving for retirement. I have no investments yet. Where do I even start?''
\item ``I'm trying to learn to play guitar as an adult. I've been at it for two months and feel like I'm not progressing. Is this normal?''
\item ``Hi there, I want to start a vegetable garden in my backyard. Small space, gets afternoon sun. What's easy to grow for a beginner?''
\item ``I just adopted an older cat from a shelter and she's been hiding under the bed for three days. Should I be worried?''
\item ``Hi, I'm thinking about getting into running. I'm completely out of shape. Should I try couch-to-5k or something else?''
\item ``I'm 35 and want to learn to swim properly. I can sort of doggy paddle. What's the best way for an adult to learn?''
\item ``Hi, my houseplant's leaves are turning yellow. I water it once a week. Did I do something wrong?''
\item ``I'm planning to propose to my partner next year and I have no idea where to start with rings. Any general advice?''
\item ``Hi, I want to start journaling but every time I try I just stare at a blank page. How do people actually keep a journal?''
\item ``I'm thinking about adopting a second dog. I already have a 5-year-old male. Any thoughts on whether this is a good idea?''
\item ``Hi, I'm 28 and have never voted before. The election is coming up and I want to participate. How do I register and vote?''
\item ``I just got my first slow cooker as a gift. What are some easy meals a beginner can try?''
\item ``Hi, my coworker recommended I read more fiction but I haven't read a novel since college. Where should I start?''
\item ``I'm trying to drink less coffee. I'm at 4 cups a day. Any tips for cutting back without total misery?''
\item ``Hi, my apartment building allows rooftop gardens but I've never grown anything. What's the absolute easiest plant to start with?''
\item ``I'm thinking about learning a second language. I'm an English speaker. Spanish or French --- which is more useful for travel?''
\item ``Hi, I want to start meditating but I keep falling asleep. Is that normal? Any tips?''
\item ``I just got a bike for commuting. The route to my office is mostly bike lanes but crosses one busy street. Any safety tips for a new commuter?''
\item ``Hi, I'm 31 and never learned to cook a steak. I bought a nice ribeye for the weekend and don't want to ruin it. Walk me through it?''
\item ``I want to start lifting weights at home. Limited space and budget. What's the minimum equipment that gets results?''
\item ``Hi, my sourdough loaf came out dense and gummy. I followed the recipe. What might have gone wrong?''
\item ``I'm planning a small dinner party for six people next weekend. I'm an okay cook but never hosted before. What do I need to know?''
\item ``Hi, I want to start hiking on weekends. Live near a national park. Never hiked before. What should I prepare for my first easy trail?''
\item ``I'm 40 and want to learn to draw. I always thought I had no talent. Is it actually possible to start as a beginner at this age?''
\item ``Hi, I just bought a film camera at a thrift store. Never shot film before. How do I figure out how to load and use it?''
\end{enumerate}

\subsection{Qwen layer-selection pilot}
\label{app:qwen-layer-selection}

We piloted four candidate Qwen3-8B layers
$\{10, 18, 23, 31\} \approx \{28\%, 50\%, 64\%, 86\%\}$ of total
depth, matched to the four Gemma 3 4B layer fractions ($27\%, 50\%,
65\%, 85\%$). For each candidate we measured relative $L_2$
reconstruction error of the Qwen Scope SAE on the residual stream
across the 60 NF prompts. L31 returned the lowest reconstruction
error of the four candidates and was selected as the analyzed layer
for the cross-family validation. The reconstruction error at L31 on
the L0\_100 Qwen-Scope variant is ${\sim}34\%$ (median $34.4\%$,
mean $37.2\%$). Control checks confirm this is a TopK-sparsity
property rather than a chat-template / base-checkpoint distribution
mismatch: errors are identical on chat-templated and raw-text
inputs, with and without subtraction of the SAE decoder bias
$b_\text{dec}$.

\section{Behavioral stratification}

\subsection{Acuity stratification of behavioral accuracy}
\label{app:acuity-stratification}

Section~\ref{sec:phase0_5_cells} reports condition-level accuracy
pooled across all 60 cases. Table~\ref{tab:acuity-stratification}
disaggregates by the corpus's acuity tagging (Low / Intermediate /
High); Table~\ref{tab:acuity-by-gold} disaggregates by the gold
triage tier.

\begin{table*}[ht]
\centering
\small
\begin{tabular}{lrccc}
\toprule
Acuity & $n$ & SL & NL & NF \\
\midrule
\multicolumn{5}{l}{\textbf{Gemma 3 4B IT}} \\
Low          & 18 & $50\%$ & $56\%$ & $44\%$ \\
Intermediate & 20 & $85\%$ & $75\%$ & $95\%$ \\
High         & 22 & $41\%$ & $36\%$ & $73\%$ \\
\midrule
\multicolumn{5}{l}{\textbf{Gemma 3 12B IT}} \\
Low          & 18 & $67\%$ & $67\%$ & $61\%$ \\
Intermediate & 20 & $80\%$ & $90\%$ & $85\%$ \\
High         & 22 & $95\%$ & $86\%$ & $68\%$ \\
\bottomrule
\end{tabular}
\caption{Per-acuity accuracy under the corpus's three-level acuity
tagging. NF is scored as both-judges-correct under 4-way
adjudication. At 4B the NF gain over NL is concentrated in the
Intermediate ($+20$pp) and High ($+37$pp) tiers and absent on Low.
At 12B the NF-below-NL pattern is concentrated on High-acuity
cases ($-18$pp), where the gold tier is C/D or D and NF's
adjacent-letter miscalibrations (Section~\ref{sec:deferred_class})
tend to land one step on the wrong side.}
\label{tab:acuity-stratification}
\end{table*}

\begin{table*}[ht]
\centering
\small
\begin{tabular}{lrccc}
\toprule
Gold & $n$ & SL & NL & NF \\
\midrule
\multicolumn{5}{l}{\textbf{Gemma 3 4B IT}} \\
A     & 8  & $12\%$ & $12\%$ & $0\%$   \\
B     & 8  & $88\%$ & $88\%$ & $75\%$  \\
C     & 10 & $30\%$ & $10\%$ & $80\%$  \\
D     & 4  & $0\%$  & $0\%$  & $0\%$   \\
A/B   & 2  & $50\%$ & $100\%$& $100\%$ \\
B/C   & 4  & $100\%$& $100\%$& $100\%$ \\
C/D   & 24 & $79\%$ & $75\%$ & $96\%$  \\
\midrule
\multicolumn{5}{l}{\textbf{Gemma 3 12B IT}} \\
A     & 8  & $38\%$ & $25\%$ & $25\%$  \\
B     & 8  & $88\%$ & $100\%$& $88\%$  \\
C     & 10 & $90\%$ & $90\%$ & $70\%$  \\
D     & 4  & $75\%$ & $50\%$ & $25\%$  \\
A/B   & 2  & $100\%$& $100\%$& $100\%$ \\
B/C   & 4  & $50\%$ & $100\%$& $100\%$ \\
C/D   & 24 & $96\%$ & $92\%$ & $83\%$  \\
\bottomrule
\end{tabular}
\caption{Per-gold-tier accuracy. The 4B NF advantage over NL is
driven by gold C ($+70$pp on NF) and gold C/D ($+21$pp on NF), the
two tiers requiring discrimination between adjacent dispositions.
The 4B singleton-A and singleton-D tiers are at floor under all
three formats. At 12B, NF underperforms NL most on gold D ($-25$pp)
and gold C ($-20$pp), where the model's tiered conditional advice
(Section~\ref{sec:deferred_class}) is read by the 4-letter judges
as the wrong middle tier.}
\label{tab:acuity-by-gold}
\end{table*}

\begin{table*}[ht]
\centering
\small
\begin{tabular}{lrccccc}
\toprule
Gold (most urgent) & $n$ & SL & NL & NF (both) & DEFERRED \\
\midrule
A (monitor)        & 8  & $75\%$ & $62\%$ & $38\%$ & $0\%$  \\
B (next weeks)     & 10 & $60\%$ & $70\%$ & $80\%$ & $10\%$ \\
C (24--48h)        & 14 & $71\%$ & $86\%$ & $64\%$ & $0\%$  \\
D (ER)             & 28 & $82\%$ & $75\%$ & $75\%$ & $4\%$  \\
\bottomrule
\end{tabular}
\caption{Qwen3-8B per-gold-tier accuracy, with gold collapsed to
the most-urgent letter (dual labels A/B$\to$B, B/C$\to$C,
C/D$\to$D). NF is worst on low-acuity (A) cases at $38\%$ ---
substantially below NL ($62\%$) and SL ($75\%$) --- consistent with
the 12B pattern of deferral leaking urgency upward.
\textsc{deferred} is the either-judge five-way rate within tier; the
$2/60$ unanimous Qwen deferrals are F15 (gold C/D, hypertensive
symptoms; collapses to D) and F19 (gold B, abnormal blood-test
report).}
\label{tab:acuity-qwen}
\end{table*}

\subsection{Triage error direction}
\label{app:triage-direction}

Under-triage is a committed prediction below the least urgent
acceptable gold tier; over-triage is above the most urgent tier.
Table~\ref{tab:triage-direction} reports the complete counts used
in Section~\ref{sec:phase0_5_cells}.

\begin{table*}[ht]
\centering
\scriptsize
\setlength{\tabcolsep}{3.5pt}
\begin{tabular}{llrrrr}
\toprule
Model & Format & Correct & Under & Over & No commit \\
\midrule
\multirow{3}{*}{4B}
 & SL & 35 & 16 & 9 & 0 \\
 & NL & 33 & 20 & 7 & 0 \\
 & NF & 43 & 5 & 8 & 4 \\
\midrule
\multirow{3}{*}{12B}
 & SL & 49 & 3 & 8 & 0 \\
 & NL & 49 & 4 & 7 & 0 \\
 & NF & 43 & 8 & 9 & 0 \\
\midrule
\multirow{3}{*}{Qwen}
 & SL & 45 & 10 & 5 & 0 \\
 & NL & 45 & 12 & 3 & 0 \\
 & NF & 37 & 8 & 6 & 9 \\
\bottomrule
\end{tabular}
\caption{Correct/under-triage/over-triage/no-commit counts out of
$60$. Multiple-choice errors skew toward over-triage at 12B, but
toward under-triage at 4B and Qwen.}
\label{tab:triage-direction}
\end{table*}

\section{Token-level activations and feature sensitivity}

\subsection{Per-token activations on all 60 cases}
\label{app:per_token}

Table~\ref{tab:per_token_full} reports per-token max activations on
the three Gemma 3 4B IT L29 medical features ($f_1=12570$,
$f_2=893$, $f_3=12845$) for all 60 cases. Across the 60 cases, the
three features fire above the JumpReLU threshold under
both NL and NF on 30, 44, and 26 cases for $f_1$, $f_2$, $f_3$
respectively. When both fire, the per-token max delta
$|a_\text{NF} - a_\text{NL}| / \max(a_\text{NF}, a_\text{NL})$ is
$\leq 8.2\%$ ($f_1$, median $0.9\%$), $\leq 4.9\%$ ($f_2$, median
$0.4\%$), and $\leq 2.1\%$ ($f_3$, median $0.6\%$). A small number
of low-magnitude near-threshold cases (1 for $f_1$, 3 for $f_2$, 2
for $f_3$; max activation $548$--$706$ across the SAE's
${\sim}3{,}500$ saturation range) fire under NL but stay below
threshold under NF; the inverse (NF-only firing) does not occur.

\begin{table*}[h]
\centering
\scriptsize
\begin{tabular}{llrrrrrr}
\toprule
Case & Gold & NL max $f_1$ & NL max $f_2$ & NL max $f_3$ & NF max $f_1$ & NF max $f_2$ & NF max $f_3$ \\
\midrule
E1 & C/D & 1\,335 & 1\,158 & 0 & 1\,348 & 1\,163 & 0 \\
E2 & B/C & 706 & 3\,398 & 1\,255 & 0 & 3\,390 & 1\,264 \\
E3 & C & 0 & 702 & 883 & 0 & 700 & 900 \\
E4 & C & 975 & 3\,423 & 2\,341 & 975 & 3\,434 & 2\,380 \\
E5 & A & 885 & 3\,342 & 882 & 903 & 3\,340 & 892 \\
E6 & B/C & 942 & 1\,209 & 988 & 941 & 1\,208 & 976 \\
E7 & C/D & 2\,573 & 3\,494 & 1\,200 & 2\,575 & 3\,482 & 1\,208 \\
E8 & A & 710 & 656 & 0 & 707 & 675 & 0 \\
E9 & D & 1\,055 & 3\,216 & 3\,001 & 1\,066 & 3\,228 & 3\,017 \\
E10 & C/D & 0 & 0 & 681 & 0 & 0 & 0 \\
E11 & C/D & 729 & 674 & 0 & 721 & 686 & 0 \\
E12 & C/D & 1\,489 & 2\,925 & 2\,214 & 1\,489 & 2\,925 & 2\,214 \\
E13 & D & 0 & 669 & 0 & 0 & 669 & 0 \\
E14 & C/D & 2\,066 & 3\,277 & 2\,014 & 2\,081 & 3\,285 & 2\,026 \\
E15 & C/D & 738 & 630 & 0 & 719 & 609 & 0 \\
E16 & C/D & 1\,204 & 1\,188 & 0 & 1\,209 & 1\,171 & 0 \\
E17 & A & 0 & 3\,644 & 0 & 0 & 3\,653 & 0 \\
E18 & B & 0 & 1\,131 & 0 & 0 & 1\,160 & 0 \\
E19 & B & 0 & 3\,016 & 0 & 0 & 3\,008 & 0 \\
E20 & A & 0 & 1\,103 & 0 & 0 & 1\,135 & 0 \\
E21 & B & 0 & 2\,962 & 0 & 0 & 3\,012 & 0 \\
E22 & C/D & 1\,004 & 3\,103 & 0 & 989 & 3\,102 & 0 \\
E23 & A/B & 0 & 3\,614 & 0 & 0 & 3\,617 & 0 \\
E24 & B & 1\,256 & 1\,063 & 2\,329 & 1\,249 & 1\,046 & 2\,335 \\
E25 & C & 3\,051 & 3\,107 & 2\,247 & 3\,050 & 3\,116 & 2\,248 \\
E26 & C/D & 1\,784 & 3\,568 & 1\,763 & 1\,811 & 3\,575 & 1\,759 \\
E27 & C/D & 1\,665 & 3\,765 & 2\,333 & 1\,647 & 3\,780 & 2\,328 \\
F1 & C/D & 1\,102 & 1\,064 & 667 & 1\,101 & 1\,070 & 0 \\
F2 & B/C & 0 & 860 & 0 & 0 & 859 & 0 \\
F3 & C & 0 & 0 & 1\,031 & 0 & 0 & 1\,029 \\
F4 & C & 1\,644 & 904 & 1\,394 & 1\,650 & 941 & 1\,409 \\
F5 & A & 746 & 0 & 706 & 741 & 0 & 710 \\
F6 & B/C & 924 & 564 & 928 & 1\,006 & 0 & 937 \\
F7 & C/D & 1\,870 & 0 & 879 & 1\,846 & 0 & 871 \\
F8 & A & 0 & 721 & 707 & 0 & 702 & 692 \\
F9 & D & 0 & 0 & 0 & 0 & 0 & 0 \\
F10 & C/D & 0 & 0 & 1\,313 & 0 & 0 & 1\,303 \\
F11 & C/D & 0 & 0 & 0 & 0 & 0 & 0 \\
F12 & C/D & 1\,266 & 914 & 1\,148 & 1\,252 & 916 & 1\,147 \\
F13 & D & 0 & 639 & 0 & 0 & 651 & 0 \\
F14 & C/D & 1\,651 & 0 & 1\,699 & 1\,662 & 0 & 1\,687 \\
F15 & C/D & 0 & 0 & 0 & 0 & 0 & 0 \\
F16 & C/D & 1\,273 & 1\,008 & 1\,123 & 1\,254 & 1\,018 & 1\,122 \\
F17 & A & 0 & 697 & 0 & 0 & 677 & 0 \\
F18 & B & 0 & 548 & 0 & 0 & 0 & 0 \\
F19 & B & 0 & 993 & 0 & 0 & 987 & 0 \\
F20 & A & 0 & 982 & 0 & 0 & 982 & 0 \\
F21 & B & 769 & 860 & 0 & 791 & 904 & 0 \\
F22 & C/D & 0 & 1\,365 & 0 & 0 & 1\,365 & 0 \\
F23 & A/B & 0 & 562 & 0 & 0 & 0 & 0 \\
F24 & B & 0 & 766 & 0 & 0 & 736 & 0 \\
F25 & C & 1\,738 & 0 & 1\,511 & 1\,751 & 0 & 1\,511 \\
F26 & C/D & 1\,340 & 1\,037 & 2\,283 & 1\,337 & 1\,044 & 2\,283 \\
F27 & C/D & 1\,498 & 1\,662 & 1\,389 & 1\,465 & 1\,725 & 1\,377 \\
MH1 & C & 737 & 3\,033 & 0 & 727 & 3\,036 & 0 \\
MH2 & C/D & 0 & 3\,004 & 0 & 0 & 3\,016 & 0 \\
MH3 & C & 0 & 3\,075 & 0 & 0 & 3\,058 & 0 \\
NH1 & C & 0 & 0 & 0 & 0 & 0 & 0 \\
NH2 & C/D & 0 & 0 & 0 & 0 & 0 & 0 \\
NH3 & C & 0 & 0 & 0 & 0 & 0 & 0 \\
\bottomrule
\end{tabular}
\caption{Per-case max activations of the three Gemma 3 4B IT L29
medical features under NL and NF for all 60 cases. Activations are
in raw SAE units (the SAE saturates near $\sim$3{,}500 on the
strongest clinical-lexicon tokens). Cells reading $0$ are
below-threshold under the JumpReLU; the corpus's NH cases (NH1,
NH2, NH3) are corpus-edge cases that fire below threshold on all
three features under both conditions.}
\label{tab:per_token_full}
\end{table*}

\subsection{Sensitivity to medical feature-set size $K$}
\label{app:phase1b_sensitivity}

The main-text mechanistic-invariance result
(Section~\ref{sec:phase1b_strata}) uses 3 medical features per
(model, layer). To check that the medical-random sMAPE gap is
not sensitive to that specific small subset, we re-run the
mechanistic-invariance test at the representative late layers
(Gemma 3 4B IT L29 and Gemma 3 12B IT L31) with $K \in \{3, 5, 10,
20\}$ medical features drawn from the same contrastive
identification ranking, against magnitude-matched random baselines.

Identical to Section~\ref{sec:mech-invariance} except we vary the
number of medical features included in the mean sMAPE. For each $K$
we take the top-$K$ entries in the contrastive ranking, max-pool
feature activations over content tokens, and compute the per-(case,
feature) sMAPE. Bootstrap mean and 95\% CI from 1{,}000 case
resamples.

\begin{table*}[ht]
\centering
\footnotesize
\setlength{\tabcolsep}{3pt}
\begin{tabular}{llccc}
\toprule
Model & $K$ & med.\ mean & rand.\ mean & paired $\Delta$ [95\% CI] \\
\midrule
4B L29  & 3  & 0.006 & 0.031 & $-0.025$ $[-0.038, -0.013]$ \\
4B L29  & 5  & 0.012 & 0.229 & $-0.217$ $[-0.268, -0.168]$ \\
4B L29  & 10 & 0.133 & 0.260 & $-0.127$ $[-0.163, -0.094]$ \\
4B L29  & 20 & 0.153 & 0.214 & $-0.061$ $[-0.089, -0.034]$ \\
\midrule
12B L31 & 3  & 0.006 & 0.741 & $-0.736$ $[-0.825, -0.643]$ \\
12B L31 & 5  & 0.005 & 0.455 & $-0.450$ $[-0.506, -0.392]$ \\
12B L31 & 10 & 0.039 & 0.592 & $-0.553$ $[-0.600, -0.507]$ \\
12B L31 & 20 & 0.087 & 0.412 & $-0.325$ $[-0.363, -0.291]$ \\
\midrule
Qwen L31 & 3 & 0.035 & 0.058 & $-0.023$ $[-0.072, +0.024]$ (ns) \\
\bottomrule
\end{tabular}
\caption{Max-pool sMAPE averaged across the top-$K$ medical
features and the 30 magnitude-matched random features, with paired
(medical $-$ random) bootstrap $\Delta$ from $1{,}000$ case
resamples. All eight Gemma cells have CIs strictly below zero. The
Qwen K=3 row matches the 3-feature set used in the main text; a
Qwen top-20 contrastive identification at L31 was not run and is
flagged as future work.}
\label{tab:phase1b_sensitivity}
\end{table*}

The medical-random gap is robust across feature-set sizes at the
two Gemma late layers. At 4B L29 the paired $\Delta$ ranges from
$-0.025$ ($K{=}3$) to $-0.217$ ($K{=}5$), narrowing to $-0.061$ at
$K{=}20$; at 12B L31 the gap stays in $[-0.74, -0.32]$ across all
$K$. Medical mean drifts upward with $K$ (4B: $0.006$ to $0.153$;
12B: $0.006$ to $0.087$), indicating that lower-ranked medical
features are somewhat noisier than the top-3 but they remain
significantly below the random baseline at every $K$ tested. The
main-text $K{=}3$ result is therefore representative of the
contrastive-ranked medical population rather than cherry-picked
from a small subset.

The medical-mean upward drift with $K$ is itself informative: it is
consistent with the SAE-as-monosemantic-probe view adopted in
Section~\ref{sec:medical-feature-id}. The top contrastive-ranked
features are the most cleanly medical-selective and respond most
similarly across formats; the broader medical-firing population in the SAE
basis includes features that are medical-content-correlated but
more sensitive to surface format. The headline interpretation (``a
small number of monosemantic medical features respond similarly to
clinical content across formats'') is what the sensitivity check quantifies, not
contradicts.

\section{Adjudication and clinical calibration}

\subsection{Five-way LLM-judge adjudication}
\label{app:deferred_adjudication}

\paragraph{Setup.} On the NF responses, we ran the two LLM judges
(\texttt{gpt-5.2-thinking-high} and \texttt{claude-sonnet-4.6})
with the four-letter adjudication prompt of Section~\ref{sec:scoring}
extended to permit a fifth output token, \textsc{deferred}. The
judges' decision rules and A/B/C/D definitions are unchanged from
the four-letter prompt; we add one rule and one category.

The \textsc{deferred} category, in the judges' own prompt:
\begin{quote}\small
The model's final recommendation is contingent on additional
information it asks the patient to provide; no single triage urgency
is committed to. Use \textsc{deferred} ONLY when the model lists
possible urgencies without committing (``could be A, could be C,
depends on Y'') and explicitly asks the patient to answer Y before
deciding, OR when the model refuses to give a triage timeframe and
only recommends ``speak to a doctor'' without specifying when.
\end{quote}

\paragraph{Inter-judge agreement on the five-way label} is $91.7\%$
at 4B (same-letter agreement, including \textsc{deferred}) and
$95\%$ at 12B ($57/60$ cases identical). On the \textsc{deferred}
class specifically, both judges agree on the same set of 4 cases at
12B (F15, F19, F23, F24); the fifth flagged \textsc{deferred} case
(F7) is flagged only by GPT, with Claude marking it as A.

\paragraph{Per-case quotes from the deferred set} (12B, both judges):

\begin{description}\small
\item[F15 (gold C/D, hypertensive symptoms)] ``If your reading is
over 180/120, go to the ER now; same- or next-day clinic if
140--180; within 1--2 weeks if below that. Please share your actual
reading.''
\item[F19 (gold B, abnormal blood-test report)] ``The urgency really
depends on the specific results. If the values are mildly abnormal,
3--6 months; moderately abnormal, 2--4 weeks; markedly abnormal,
within 1 week.''
\item[F23 (gold A/B, post-viral leukopenia)] Tiered framework over
$\{1$--$2$ weeks, within 1 week, within 2--3 days, emergency$\}$
depending on the actual count.
\item[F24 (gold B, lab follow-up)] Three-tier conditional advice
(2--4 weeks / 1--2 weeks / immediate) conditional on the actual
elevation.
\end{description}

\paragraph{Where the deferred cases came from under four-letter
adjudication.} Under the original four-letter prompt, all four cases
were labeled (B, B, B, B) by both judges. The B label is plausible
by default (``see a doctor in the next few weeks'') but does not
reflect the model's central reasoning, which is a tiered
conditional. Under the five-letter prompt, both judges promote
these to \textsc{deferred}, which both reduces 12B's accuracy on the
conventional four-letter score (GPT $71.7\% \rightarrow 65.0\%$,
Claude $71.7\% \rightarrow 66.7\%$) and yields a more faithful
description of the model's actual behavior.

\subsection{Clinician adjudication}
\label{app:clinician}

\paragraph{Protocol.} A clinician co-author (Jialei Sheng, NSW Health)
reviewed a blinded
sample of $n=16$ cases stratified across
four cells of 4 each, drawn from the 60-case corpus:
\texttt{12B\_both\_deferred} ($n=4$, model defers tiered-conditional
advice), \texttt{12B\_clear\_correct} ($n=4$, both LLM judges agree
and match gold), \texttt{12B\_clear\_incorrect} ($n=4$, both judges
agree but not in gold), and \texttt{4B\_format\_flipped} ($n=4$, NL
letter wrong, NF judged correct by both LLMs). The clinician saw
only the patient message and the model's free-text response; both
gold labels and LLM-judge labels were withheld. For each case the
clinician rated: (1) primary triage on the five-class A/B/C/D
/DEFERRED scale, using the same prompt the LLM judges used; (2)
deferral-appropriateness (yes/no/NA) if the chosen label was
DEFERRED; (3) confidence (1--5); (4) whether they agreed with the
model's recommendation (yes/partial/no); (5) free-text notes.

\paragraph{Headline results.}

\begin{table*}[ht]
\centering
\small
\begin{tabular}{lr}
\toprule
Metric & Value \\
\midrule
LLM-judge vs.\ clinician triage agreement & \textbf{14/16 (87.5\%)} \\
Cohen's $\kappa$ (5-class)                & \textbf{0.808} \\
Model deferrals judged clinically appropriate & $3/4$ (75\%) \\
Clinician fully agrees with model recommendation & $8/16$ (50\%) \\
Clinician partially agrees with model     & $8/16$ (50\%) \\
Clinician fully disagrees with model      & $\mathbf{0/16}$ \textbf{(0\%)} \\
Clinician deferral rate (this 16-case sample) & $6/16$ (37.5\%) \\
Model deferral rate (full 60-case corpus) & $4/60$ (6.7\%) \\
\bottomrule
\end{tabular}
\caption{Clinician adjudication summary on the $n=16$ stratified
sample. Cohen's $\kappa = 0.808$ between the LLM-judge majority
verdict and the clinician's primary triage label, on the five-class
$\{A, B, C, D, \textsc{deferred}\}$ scale.}
\label{tab:clinician}
\end{table*}

\paragraph{Two cases where the clinician diverges from the LLM
judges.}

\begin{description}\small
\item[R13 (case E5, gold A)] Both LLM judges and the model committed
to A (matching gold); the clinician marked DEFERRED with the
deferral judged clinically appropriate, on the grounds that the case
``requires a time element to manifest diagnosis vs.\ subsiding
symptoms''. This is a case where the model committed to the
gold-correct letter but the clinician argues the commitment was
premature.
\item[R06 (case F9, gold D)] Both LLM judges read the model as
recommending C (24--48 hours); the clinician marked DEFERRED with
the deferral judged inappropriate, noting that an urgent
recommendation was warranted. The inverse pattern: the model's
hedging was read as deferral by the clinician, but the clinician
thinks the clinical situation called for a definite urgent
recommendation.
\end{description}

\paragraph{Reading.} Agreement with the clinician is high on this
enriched sample ($\kappa=0.808$), providing face-validity calibration
for the LLM-judge labels. Three of four 12B tiered deferrals are
judged clinically appropriate. The clinician defers more often on
the sampled cases ($6/16$) than the model does on the full corpus
($4/60$), and records no full disagreement with a model response.
Because the sample is small and deliberately stratified, these
observations do not estimate deployment safety or population
deferral prevalence.

\section{Verification and full invariance results}

\subsection{Verified baseline numerics}
\label{app:verified_baselines}

A small number of numerical claims about SAE properties and
intervention magnitudes appear in the main text without their
derivation. We list the verified source values here, recomputed
end-to-end from the same checkpoints used for the analyses.

\begin{table*}[ht]
\centering
\scriptsize
\begin{tabular}{p{4.9cm} l l}
\toprule
Claim & Reported & Verified \\
\midrule
Gemma 3 4B IT L29 per-token $L_0$ (active features), $n{=}3950$ tokens & ``${\sim}35$--$90$ active (median $57$)'' & median 57.0, mean 58.2, $[$p5, p95$] = [$36, 86$]$ \\
Gemma 3 4B IT L29 per-token residual norm, same sample & ``${\sim}60{,}000$ per token'' & mean 60{,}583, median 59{,}433 \\
Intervention: mean norm subtracted per token (case E1, NL) & 264 & 264.4 \\
Intervention: peak norm subtracted on a single token & 6{,}795 & 6{,}799.7 \\
Intervention: mean contribution as \% of residual norm & ${\sim}0.4\%$ & 0.44\% \\
Intervention: peak contribution as \% of residual norm & ${\sim}11\%$ & 10.97\% \\
Gemma Scope 2 4B L29 relative $L_2$ reconstruction error & ${\sim}14\%$ & B\_mean 14.0\%, D\_mean 13.6\% \\
Qwen-Scope L31 relative $L_2$ reconstruction error (L0\_100), $n{=}1663$ tokens & ${\sim}34\%$ & median 34.4\%, mean 37.2\% \\
ActAdd: $\alpha \cdot \lVert v \rVert / \lVert r \rVert$ at $\alpha{=}4$ & ${\sim}6.7\%$ & $4 \cdot 1012.66 / 60{,}583 = 6.69\%$ \\
\bottomrule
\end{tabular}
\caption{Verified numerical claims, recomputed end-to-end.}
\label{tab:verified_baselines}
\end{table*}

\subsection{Full per-layer / per-stratum tables}
\label{app:full-tables}

The headline table (Table~\ref{tab:phase1b_headline}) reports only
the late layer per model. Tables~\ref{tab:full_4b},
\ref{tab:full_12b}, and \ref{tab:full_restricted} give the full
per-layer / per-stratum sweeps and the restricted-random-pool
comparison, both under max-pool aggregation with sMAPE and cosine
side-by-side. Cells with $n_\text{cos}$ noted in parentheses have
effective cosine $n < n_\text{sMAPE}$ because one of the
medical/random feature subvectors is identically zero for some
cases (undefined cosine).

\begin{table*}[ht]
\centering
\footnotesize
\setlength{\tabcolsep}{3pt}
\begin{tabular}{rlrcc}
\toprule
L & Stratum & $n$ & $\Delta$sMAPE [95\% CI] & $\Delta$cos [95\% CI] \\
\midrule
\multirow{5}{*}{9}
  & both-right     & 29 & $-0.232$ $[-0.265,-0.201]$ & $+0.023$ $[+0.019,+0.026]$ \\
  & both-wrong     & 12 & $-0.202$ $[-0.238,-0.167]$ & $+0.024$ $[+0.019,+0.029]$ \\
  & NF-only-right & 14 & $-0.213$ $[-0.241,-0.186]$ & $+0.024$ $[+0.016,+0.034]$ \\
  & NL-only-right & 1  & $-0.197$ (n=1)             & $+0.019$ (n=1) \\
  & judges\_dis.    & 4  & $-0.207$ $[-0.229,-0.190]$ & $+0.026$ $[+0.020,+0.033]$ \\
\midrule
\multirow{5}{*}{17}
  & both-right     & 29 & $-0.269$ $[-0.305,-0.231]$ & $+0.035$ $[+0.023,+0.050]$ ($n_c{=}28$) \\
  & both-wrong     & 12 & $-0.290$ $[-0.363,-0.225]$ & $+0.039$ $[+0.025,+0.056]$ \\
  & NF-only-right & 14 & $-0.258$ $[-0.315,-0.205]$ & $+0.027$ $[+0.017,+0.039]$ ($n_c{=}13$) \\
  & NL-only-right & 1  & $-0.206$ (n=1)             & $+0.047$ (n=1) \\
  & judges\_dis.    & 4  & $-0.209$ $[-0.269,-0.148]$ & $+0.024$ $[+0.012,+0.041]$ \\
\midrule
\multirow{5}{*}{22}
  & both-right     & 29 & $-0.090$ $[-0.142,-0.023]$ & $+0.037$ $[+0.007,+0.061]$ \\
  & both-wrong     & 12 & $-0.093$ $[-0.175,+0.010]$ & $+0.022$ $[-0.030,+0.060]$ \\
  & NF-only-right & 14 & $-0.141$ $[-0.186,-0.095]$ & $+0.043$ $[+0.026,+0.058]$ \\
  & NL-only-right & 1  & $-0.170$ (n=1)             & $+0.050$ (n=1) \\
  & judges\_dis.    & 4  & $-0.141$ $[-0.211,-0.052]$ & $+0.041$ $[-0.006,+0.070]$ \\
\midrule
\multirow{5}{*}{29}
  & both-right     & 29 & $-0.275$ $[-0.329,-0.229]$ & $+0.054$ $[+0.039,+0.072]$ ($n_c{=}24$) \\
  & both-wrong     & 12 & $-0.336$ $[-0.409,-0.265]$ & $+0.060$ $[+0.025,+0.101]$ ($n_c{=}11$) \\
  & NF-only-right & 14 & $-0.338$ $[-0.402,-0.280]$ & $+0.049$ $[+0.028,+0.071]$ ($n_c{=}12$) \\
  & NL-only-right & 1  & $-0.334$ (n=1)             & $+0.058$ (n=1) \\
  & judges\_dis.    & 4  & $-0.274$ $[-0.331,-0.217]$ & $+0.075$ $[+0.031,+0.139]$ \\
\bottomrule
\end{tabular}
\caption{Gemma 3 4B IT --- max-pool sMAPE and cosine per stratum at
all four analyzed layers. The single NL-only-right case
is reported for completeness but excluded from headline claims.}
\label{tab:full_4b}
\end{table*}

\begin{table*}[ht]
\centering
\footnotesize
\setlength{\tabcolsep}{3pt}
\begin{tabular}{rlrcc}
\toprule
L & Stratum & $n$ & $\Delta$sMAPE [95\% CI] & $\Delta$cos [95\% CI] \\
\midrule
\multirow{3}{*}{12}
  & both-right     & 43 & $+0.009$ $[-0.026,+0.051]$ & $-0.002$ $[-0.012,+0.004]$ \\
  & both-wrong     & 11 & $+0.069$ $[-0.048,+0.222]$ & $-0.020$ $[-0.067,+0.005]$ \\
  & NL-only-right & 6  & $+0.020$ $[-0.125,+0.233]$ & $+0.009$ $[+0.001,+0.020]$ \\
\midrule
\multirow{3}{*}{24}
  & both-right     & 43 & $+0.233$ $[+0.183,+0.292]$ & $-0.100$ $[-0.153,-0.053]$ \\
  & both-wrong     & 11 & $+0.366$ $[+0.185,+0.589]$ & $-0.110$ $[-0.183,-0.046]$ \\
  & NL-only-right & 6  & $+0.429$ $[+0.311,+0.517]$ & $-0.256$ $[-0.336,-0.154]$ \\
\midrule
\multirow{3}{*}{31}
  & both-right     & 43 & $-0.222$ $[-0.241,-0.203]$ & $+0.093$ $[+0.078,+0.108]$ \\
  & both-wrong     & 11 & $-0.232$ $[-0.277,-0.189]$ & $+0.094$ $[+0.067,+0.128]$ \\
  & NL-only-right & 6  & $-0.262$ $[-0.312,-0.220]$ & $+0.093$ $[+0.066,+0.129]$ \\
\midrule
\multirow{3}{*}{41}
  & both-right     & 43 & $-0.110$ $[-0.136,-0.086]$ & $+0.034$ $[+0.027,+0.042]$ \\
  & both-wrong     & 11 & $-0.136$ $[-0.192,-0.083]$ & $+0.035$ $[+0.023,+0.051]$ \\
  & NL-only-right & 6  & $-0.200$ $[-0.278,-0.116]$ & $+0.035$ $[+0.019,+0.055]$ \\
\bottomrule
\end{tabular}
\caption{Gemma 3 12B IT --- max-pool sMAPE and cosine per stratum at
all four analyzed layers. The sign of the effect flips between
shallow/mid layers (L12, L24, positive sMAPE / negative cosine) and
deep layers (L31, L41, negative sMAPE / positive cosine); the two
metrics agree on direction at every cell. Per-case values are from
the original 12B pipeline run; an independent v2 re-extraction
reproduces the same medical feature set ($\{3, 338, 329\}$ at L24,
$\{130, 85, 4773\}$ at L31) and every qualitative cell, with
small numerical drift ($\leq 5\%$) within bf16 inference
non-determinism across GPU instances.}
\label{tab:full_12b}
\end{table*}

\paragraph{Interpreting the 12B sign reversal.}
At L12/L24, contrastive selection preferentially retains
surface-word features that respond to medical vocabulary and are
sensitive to the answer-option text appended only in NL. At L31/L41,
the selected features are content-anchored: their peaks fall inside
the vignette in $98$--$100\%$ of case--feature pairs. Both metrics
agree within every cell, and an independent extraction reproduces
the direction, so the reversal is a stable depth effect rather than
sampling noise. Our conclusions are accordingly scoped to the late
layers where the content-anchored features and the format direction
are jointly analyzed.

\begin{table*}[ht]
\centering
\footnotesize
\setlength{\tabcolsep}{3pt}
\begin{tabular}{lrcc}
\toprule
Stratum & $n$ & $\Delta$sMAPE [95\% CI] & $\Delta$cos [95\% CI] \\
\midrule
both-right     & 29 $(n_c{=}25)$ & $-0.285$ $[-0.340,-0.213]$ & $+0.094$ $[+0.070,+0.115]$ \\
both-wrong     & 12 $(n_c{=}11)$ & $-0.329$ $[-0.367,-0.290]$ & $+0.108$ $[+0.085,+0.129]$ \\
NF-only-right  & 14 $(n_c{=}12)$ & $-0.272$ $[-0.345,-0.173]$ & $+0.096$ $[+0.071,+0.125]$ \\
NL-only-right  & 1               & $-0.337$ (n=1)             & $+0.114$ (n=1) \\
judges\_dis.    & 4               & $-0.293$ $[-0.345,-0.252]$ & $+0.092$ $[+0.070,+0.127]$ \\
\bottomrule
\end{tabular}
\caption{Restricted random pool at Gemma 3 4B IT L29
(magnitude-matched + firing on $\geq 25\%$ of the $120$
$\text{NL}\cup\text{NF}$ prompts; pool size $3{,}258$ features),
max-pool aggregation, on the canonical stratification (F1 in
NF-only-right). All populated strata's $95\%$ CIs exclude zero on
both metrics: the medical-vs-restricted-random gap survives the
firing-threshold restriction in every stratum, with modest
magnitude changes relative to the unrestricted control
(Table~\ref{tab:full_4b}). Per-case random sMAPE/cosine are
averaged across $1{,}000$ random draws of $30$ features from the
restricted pool to stabilize against single-seed sampling
variance; the per-case bootstrap CI then propagates through the
case-clustered resample ($B{=}2{,}000$).}
\label{tab:full_restricted}
\end{table*}

\begin{table*}[ht]
\centering
\footnotesize
\setlength{\tabcolsep}{3pt}
\begin{tabular}{lrcccc}
\toprule
Stratum & $n$ & med.\ sMAPE & rnd.\ sMAPE & med.\ cos & rnd.\ cos \\
\midrule
both-right      & 35 & 0.003 & 0.064 & 1.000 & 0.991 \\
NF-only-right   & 6  & 0.003 & 0.124 & 1.000 & 0.975 \\
NL-only-right   & 8  & 0.002 & 0.054 & 1.000 & 0.994 \\
both-wrong      & 6  & 0.004 & 0.042 & 1.000 & 0.994 \\
judges-disagree & 5  & 0.004 & 0.136 & 1.000 & 0.980 \\
\bottomrule
\end{tabular}
\caption{Qwen3-8B at L31 (L0\_100 Qwen-Scope variant) --- max-pool
sMAPE and cosine per stratum, joined against the Qwen NL-NF
correctness labels from the behavioral run
(Section~\ref{sec:phase0_5_cells}). Medical features stay at
near-identical activations under NL-NF across every behavioral
stratum (medical sMAPE $\leq 0.005$, cos $\geq 0.9999$), while the
magnitude-matched random control shows substantially larger
format-induced perturbation in every stratum (random sMAPE
$0.04$--$0.14$, cos $0.975$--$0.994$). The paired
medical${-}$random $\Delta$ sMAPE has a bootstrap $95\%$ CI
strictly below zero in every populated stratum, matching the
per-stratum pattern at 4B and 12B.}
\label{tab:full_qwen}
\end{table*}

\subsection{Metric consistency under aggregation}
\label{app:metric_consistency}

Section~\ref{sec:phase1b_strata} reports two summary metrics
(sMAPE and cosine) under max-pool aggregation. The choice of
aggregation matters at one cell (Qwen L31) and not at any other.
Table~\ref{tab:metric_consistency} reports the medical${-}$random
$\Delta$ for both metrics under both aggregations, per cell.

\begin{table*}[ht]
\centering
\footnotesize
\setlength{\tabcolsep}{4pt}
\begin{tabular}{llcc}
\toprule
Cell & pool & $\Delta$sMAPE & $\Delta$cos \\
\midrule
4B  L9   & mean & $-0.219$ & $+0.023$ \\
4B  L9   & max  & $-0.231$ & $+0.086$ \\
4B  L17  & mean & $-0.265$ & $+0.033$ \\
4B  L17  & max  & $-0.273$ & $+0.074$ \\
4B  L22  & mean & $-0.107$ & $+0.036$ \\
4B  L22  & max  & $-0.139$ & $+0.044$ \\
4B  L29  & mean & $-0.303$ & $+0.056$ \\
4B  L29  & max  & $-0.264$ & $+0.107$ \\
12B L12  & mean & $+0.040$ & $-0.004$ \\
12B L12  & max  & $+0.021$ & $-0.005$ \\
12B L24  & mean & $+0.172$ & $-0.093$ \\
12B L24  & max  & $+0.277$ & $-0.117$ \\
12B L31  & mean & $-0.238$ & $+0.121$ \\
12B L31  & max  & $-0.228$ & $+0.093$ \\
12B L41  & mean & $-0.103$ & $+0.070$ \\
12B L41  & max  & $-0.124$ & $+0.034$ \\
\rowcolor[gray]{0.92} Qwen L31 & mean & $-0.064$ & $\mathbf{-0.023}$ (disagree) \\
\rowcolor[gray]{0.92} Qwen L31 & max  & $-0.114$ & $+0.010$ (agree) \\
\bottomrule
\end{tabular}
\caption{Medical${-}$random $\Delta$ under mean-pool vs.\ max-pool
across all (model, layer) cells, on both sMAPE and cosine. All
cells with $|\Delta| > 0.01$ have 95\% bootstrap CIs strictly
excluding zero except 12B L12. On Gemma-Scope (JumpReLU) the two
metrics agree on direction under \emph{both} aggregations. Only
Qwen-Scope (TopK) shows aggregation-sensitivity: under mean-pool,
sMAPE reads medical-more-invariant while cosine reads
medical-less-similar; under max-pool both metrics agree
(medical-more-invariant).}
\label{tab:metric_consistency}
\end{table*}

The mean-pool disagreement at Qwen L31 is mechanistically
attributable to the TopK SAE forcing nearly all features to zero on
any given token. This dilutes per-case mean-pool feature subvectors
to near-zero in both NL and NF, exposing both metrics to noise
(denominator-clamp artifact for sMAPE; direction noise on near-zero
vectors for cosine). Max-pool aggregates the peak per-feature
firing, which is non-zero whenever the feature is in the top-$k$ on
at least one content token, and is therefore robust to TopK zeroing.
We adopt max-pool throughout this analysis for this reason and for
consistency with the direction analysis in
Section~\ref{sec:direction-test} (also max-pool).

\subsection{Pooling conventions}
\label{app:pooling-conventions}

Gemma/Qwen content-pool conventions differ for in-distribution
reasons. Gemma Scope 2 is trained on activations of the IT models
with chat-template structure, so we present chat-templated input and
pool over user-content tokens between
\texttt{<start\_of\_turn>user\textbackslash n} and
\texttt{<end\_of\_turn>}. Qwen-Scope is trained on activations of
the base (pre-IT) Qwen3-8B checkpoint, so we feed raw text and pool
over the entire input to keep the residual stream in-distribution
for the SAE. The conventions exhaust the choices the published SAE
training settings permit.

\subsection{Why three features --- methodological note}
\label{app:why-three-features}

Our methodology follows the SAE-as-monosemantic-probe
tradition~\cite{bricken2023monosemanticity,templeton2024scaling}:
SAE features are treated as individually interpretable, named units,
and claims are made about the identities of small subsets of
features rather than the predictive power of a large feature basis.
We adopt this approach because our research question concerns the
\emph{representational content} of the residual stream (``what does
the model represent and is that preserved across formats?''), not
the discriminability of medical vs.\ non-medical content via a
downstream classifier. A complementary line of work using larger
SAE-feature sets fitted to a downstream classification
head~\cite{marks2024sparse,makelov2024principled} is appropriate
for different research questions --- \emph{can SAE features
discriminate true/false statements, or category $X$ from $Y$?} ---
where the unit of analysis is the classifier and a richer basis
reduces noise. For our question, three monosemantic, contrastive-identified
features suffice as a probe; the result must hold against a properly
matched null rather than scale with feature-set size. We mitigate
the ``cherry-picked subset'' concern with two random baselines
(magnitude-matched and content-restricted), both of which yield
higher sMAPE (and lower cosine) than the medical features across
the headline (model, layer) combinations
(Section~\ref{sec:phase1b_strata}). The $K$-sweep in
Appendix~\ref{app:phase1b_sensitivity} ($K \in \{3, 5, 10, 20\}$ at
4B L29 and 12B L31) confirms the medical-random gap is stable
across feature-set size, supporting that the 3-feature result
reflects a population-level pattern rather than a small-sample
artifact.

\paragraph{Why magnitude-match the random pool.} An uncontrolled
random sample from the 16{,}384 (Gemma) or 65{,}536 (Qwen) features
would be dominated by features that rarely fire on the
60-case corpus. Their sMAPE values collapse to zero through the
$\varepsilon$ denominator floor in the sMAPE formula
(Section~\ref{sec:mech-invariance}), which biases the random mean
downward and artificially narrows the medical-random gap.
Restricting the pool to features whose mean activation falls in the
band $[0.5\min(\text{med}), 2.0\max(\text{med})]$ confines the
random control to features that fire at comparable magnitudes to
the medical features, making the comparison faithful.

\section{Interventions and format-direction evidence}

\subsection{Intervention details}
\label{app:intervention-details}

\paragraph{Discrete SAE-feature ablation.} For each NL forward pass
at Gemma 3 4B IT L29 we register a hook that subtracts the
SAE-reconstructed contribution of the three top format-direction
features (3833, 10012, 980). A control arm ablates three
magnitude-matched random features. A diagnostic confirms the hook
fires on every forward and modifies the residual at the expected
magnitude: mean 264 norm subtracted per token, peak 6{,}795 on the
strongest answer-key tokens. Relative to the L29 residual norm
(${\sim}60{,}000$ per token), this is ${\sim}0.4\%$ on average and
peaks at ${\sim}11\%$ on the strongest scaffold tokens ---
insufficient to flip the next-token argmax.

\paragraph{Continuous ActAdd steering.} Following the
activation-addition formulation of~\citet{turner2023steering}, we
compute $v = \langle r_\text{NL}\rangle - \langle r_\text{NF}\rangle$
at L29
(case-averaged mean residuals at content tokens;
$\lVert v\rVert = 1{,}012$). For each NL forward pass at
$\alpha \in \{0, 0.5, 1, 2, 4\}$ we register a hook that adds
$-\alpha \cdot v$ to the L29 output at every token. At $\alpha = 4$
the perturbation magnitude is
$4 \cdot 1012.66 / 60{,}583 = 6.69\%$ of residual norm
(Appendix~\ref{app:verified_baselines}). Only cases F2 and E6 shift
(F2 at $\alpha = 2$, E6 at $\alpha = 4$); both are gold $=$ B/C and
shift from C to B, remaining within the permissive gold range.

\subsection{Format-direction top-token examples}
\label{app:format_features}

At Gemma 3 4B IT L29 we inspect the three features most aligned
with the (NL${-}$NF) max-pool direction (3833, 10012, 980) across
all NL-NF prompts. Their top-3 activations are all NL-only literal
answer-key scaffold tokens: 3833 on ``next'' in option B, 10012 on
``the'' in option D, and 980 on answer-key ``='' syntax. The
medical features (12570, 893, 12845) instead fire on
clinical-content tokens at matched magnitudes across NL and NF.

\begin{figure*}[t]
\centering
\includegraphics[width=\linewidth]{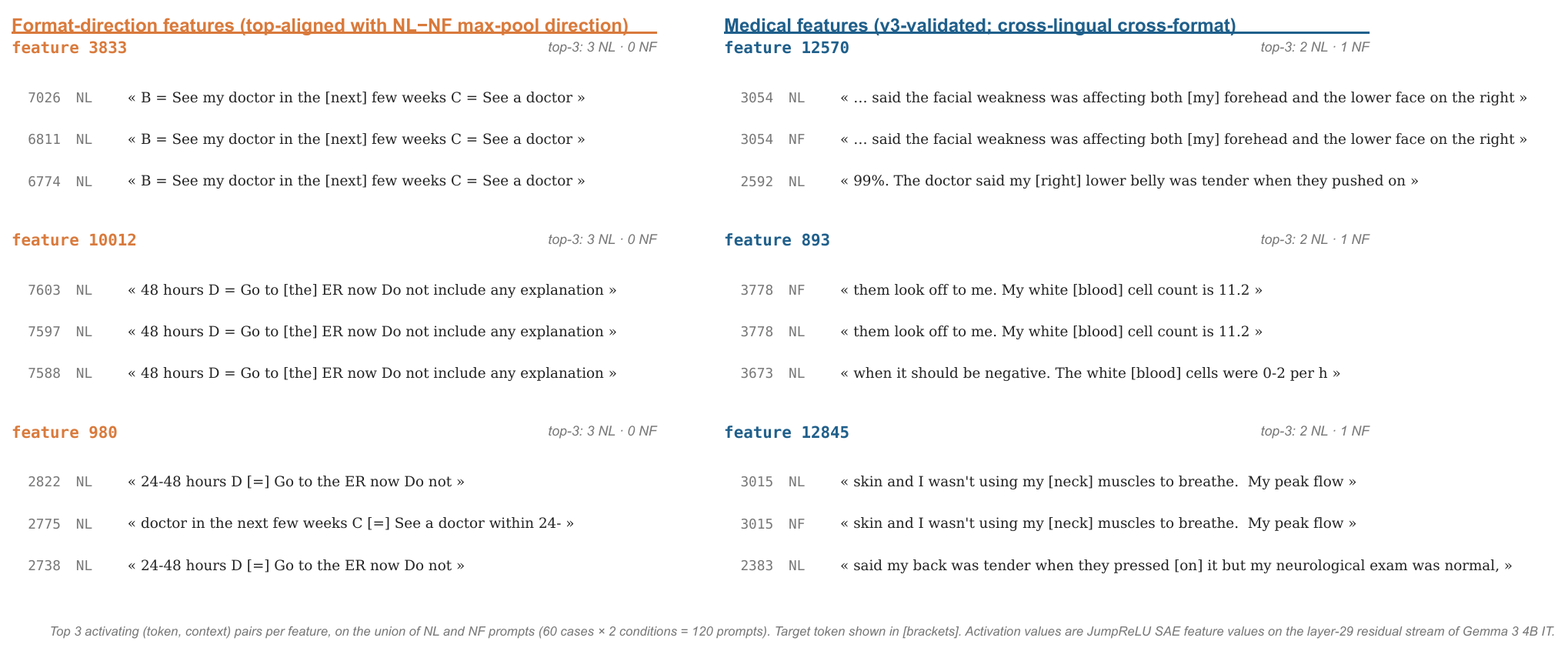}
\caption{Top-activating (token, context) pairs at Gemma 3 4B IT
L29 for the three top format-direction features (left: 3833, 10012,
980) and the three contrastive-identified medical features (right:
12570, 893, 12845), across all 60 cases $\times$ \{NL, NF\} prompts.
Format-direction features fire exclusively in NL on the literal
answer-key scaffold tokens; medical features fire on clinical-content
tokens at matched magnitudes across both conditions.}
\label{fig:format_features}
\end{figure*}

\subsection{Per-case residual-direction consistency}
\label{app:direction-consistency}

The direction analysis averages the case-level residual differences
$\Delta r_i=r_{i,\mathrm{NL}}-r_{i,\mathrm{NF}}$. To verify that
averaging does not manufacture a common direction, we compare every
$\Delta r_i$ with $\overline{\Delta r}$ at Gemma 3 4B IT L29
(mean-pooled residuals). Cosine similarity has mean $0.887$, median
$0.890$, and minimum $0.758$ across all 60 cases; the mean cosine
between pairs of individual cases is $0.789$. The aggregate
direction is therefore present in every case, not created by a
small subset.

\section{NLA and invariance checks}

\subsection{NLA verification details}
\label{app:nla-details}

\paragraph{Protocol.} For each of the $60$ cases we run two
forward passes of Gemma 3 12B IT --- one with the NL prompt, one
with the NF prompt --- and capture L32 residual-stream activations
at seven token positions, listed below:
\begin{itemize}\small
\setlength\itemsep{0pt}
\item \textsc{content}: the last clinical-content token (the
``?'' of ``how soon should I follow up?''); identical under NL
and NF, captured under both;
\item \textsc{decision}: NL-only; the last user-message token,
whose hidden state drives the first generated letter;
\item \texttt{letter\_A}, \texttt{letter\_B}, \texttt{letter\_C},
\texttt{letter\_D}: NL-only; the four answer-key letter tokens
inside the multiple-choice scaffold.
\end{itemize} This yields
$60 \times 7 = 420$ activation records. Each is fed to the NLA
actor-verbalizer under greedy decoding (temperature 0, max 200 new
tokens), with the actor prompt template released with the NLA
checkpoint, injection scale 80{,}000, and the $\sqrt{d}$
embed-scaling Gemma 3 requires. The NLA AV is trained at L32, one
layer adjacent to our L31 SAE analysis; the output-mapping result
holds at both L31 and L41 in our 12B data, so this one-layer drift
sits inside a stable encoding region.

\paragraph{Consistency check.} The \texttt{NL\_content} and
\texttt{NF\_content} L32 activations agree to median relative $L_2$
difference $0.28\%$ (maximum $0.90\%$) on residuals of norm
${\sim}60{,}000$ --- consistent with bf16 numerical noise as
expected under causal attention with byte-identical prefix tokens.
The NLA greedy outputs agree semantically (both 100\% medical) at
this position, with character-identical strings on $17/60$ cases
and minor surface variation on the rest.

\paragraph{Representative NLA outputs (case E1).}
\begin{description}\small
\item[\texttt{NF\_content}] (``?'' of ``follow up?''): \emph{``Medical
Q\&A format established: structured clinical question seeking advice,
with a patient reporting a post-surgical follow-up concern.''}
\item[NL decision] (``.'' of ``\ldots extra words.''):
\emph{``Structured quiz format with answer choices, requiring a
single lettered answer to follow the prompt about a
medical/educational passage.''}
\item[\texttt{NL\_letter\_B}]: \emph{``Structured quiz format with
numbered options and a question prompt establishes a list of answer
choices for a moral/ethical dilemma\ldots\ pattern of lettered
options, requiring a second option label to follow.''}
\end{description}

\paragraph{Per-letter pattern.} Across the four NL-only answer-key
letter tokens, all four are $60/60$ \textsc{sca-primary}. They
differ on the \textsc{medical} axis: \texttt{letter\_B} has $52/60$
\textsc{med-no} (the cleanest ``no medical content'' verdict in the
dataset) while \texttt{letter\_D} is $60/60$ \textsc{med-partial}
because ``\emph{D = Go to the ER now}'' carries strong clinical
context. This per-token variation matches the literal clinical
loading of each option's text and matches the top-token analysis in
Appendix~\ref{app:format_features}, which identifies individual SAE
features firing on these same scaffold tokens.

\subsection{Token-mask decomposition (causal-masking sanity check)}
\label{app:token_masks}

In a causal decoder transformer, hidden states at the shared
clinical-prefix tokens of NL and NF cannot differ: NL appends the
multiple-choice scaffold \emph{after} the prefix, so causal masking
forbids the appended tokens from influencing earlier states.
Pooling SAE feature activations over a span that includes those
shared tokens therefore contributes a trivially-zero sMAPE
component. We verified this prefix-identity claim by tokenizing all
$60$ paired NL-NF prompts under each model's own tokenizer: at
Gemma 3 4B IT and 12B IT, $60/60$ cases pass --- the vignette text
re-tokenizes identically inside NF and the shared prefix covers all
$255$ vignette tokens. At Qwen3-8B, $60/60$ vignettes are
internally identical, but Qwen's BPE merges the trailing ``?\textbackslash n\textbackslash n''
into one token in NL (where the scaffold follows) versus keeping
``?'' as a separate token in NF (where \texttt{<|im\_end|>}
follows); the divergence shifts one token earlier than the
vignette end. $\sim99.6\%$ of vignette positions are byte-identical
at Qwen too, and the small non-zero vignette-mask sMAPE we observe
there ($0.0018$ / $0.0031$ medical/random) is consistent with this
single-token BPE boundary effect plus bf16 noise. To make the
analysis concrete we partition the NL$\cup$NF token stream into
three masks:

\begin{description}
\setlength\itemsep{0pt}
\item[Vignette mask] tokens of the shared clinical narrative, identical
in NL and NF up to chat-template framing.
\item[Scaffold mask] NL-only tokens of the forced-letter instruction
plus the A--D answer key.
\item[Decision token] the last user-message token whose hidden state
drives the first generated letter (in NL this is the chat-template
suffix following the scaffold; in NF the closing punctuation of the
patient narrative).
\end{description}

\begin{table*}[ht]
\centering
\footnotesize
\setlength{\tabcolsep}{4pt}
\begin{tabular}{lccc}
\toprule
Model & vignette mask & full content & scaffold (NL) \\
& med / rnd & med / rnd & --- \\
\midrule
4B L29   & $0.004$/$0.006$ & $0.004$/$0.276$ & --- \\
12B L31  & $0.003$/$0.005$ & $0.003$/$0.123$ & --- \\
Qwen L31 & $0.002$/$0.003$ & $0.026$/$0.128$ & --- \\
\bottomrule
\end{tabular}
\caption{Per-mask sMAPE (medical / random max-pool) for the three
headline (model, layer) cells. On the shared vignette mask both
medical and random sMAPE collapse to $\approx 0.002$--$0.006$, as
expected under causal masking --- the vignette-mask sanity check
passes. On the full-content max-pool, medical sMAPE remains near
zero while random sMAPE rises to $0.12$--$0.28$, the headline
medical-random gap.}
\label{tab:masked_invariance}
\end{table*}

\paragraph{Medical-feature peaks are anchored in the shared
vignette.} Under max-pool aggregation, the case-level statistic is
dominated by the token at which each feature reaches its peak. For
the three medical features at each model, the peak-location
fraction inside the vignette mask is
$99.4\%/99.4\%$ (4B NL/NF), $98.3\%/100.0\%$ (12B NL/NF),
and $100.0\%/100.0\%$ (Qwen NL/NF, L0\_100); on the
structured-input pair the corresponding Qwen numbers are
$99.4\%$ (SL) and $98.8\%$ (SF). Medical-feature peaks are
anchored in the clinical narrative across all three models; the
medical-random gap on the full-content mask reflects this
anchoring, not trivial prefix overlap.

\paragraph{Decision-token caveat.} The decision-token comparison
as defined (last \emph{prompt} token) compares syntactic
chat-template suffix tokens whose feature activations are dominated
by generation-prep state; we report no decision-token sMAPE in
Table~\ref{tab:masked_invariance} for this reason. A better
analysis would compare the last \emph{content} token under each
format; we leave this redesign as future work.

\subsection{Magnitude-matched random resampling}
\label{app:resample}

The fixed-seed random control of Section~\ref{sec:phase1b_strata}
uses a single draw of 30 features from the magnitude-matched pool.
To quantify the medical-random gap's robustness to that draw,
we resample the random pool $1{,}000$ times under the same
magnitude-matching constraint
($[0.5\min(\text{med}), 2.0\max(\text{med})]$) and re-compute the
per-case sMAPE in each draw. Table~\ref{tab:resample} reports the
$5$th--$95$th percentile range of random sMAPE across draws and the
one-sided permutation $p$-value for the medical-random gap
(fraction of random-pool draws whose mean sMAPE is at or below the
observed medical mean).

\begin{table*}[ht]
\centering
\footnotesize
\setlength{\tabcolsep}{4pt}
\begin{tabular}{lccc}
\toprule
Cell & med.\ sMAPE & rnd.\ sMAPE (5--95\%) & perm.\ $p$ \\
\midrule
4B L29   & $0.004$ & $0.276$ $[0.138, 0.429]$ & $<0.001$ \\
12B L31  & $0.003$ & $0.123$ $[0.072, 0.171]$ & $<0.001$ \\
Qwen L31 & $0.002$ & $0.218$ $[0.094, 0.361]$ & $<0.001$ \\
\bottomrule
\end{tabular}
\caption{$1{,}000$-resample magnitude-matched random control with
permutation $p$-values for the medical-random sMAPE gap at the
headline layer per model. The medical-feature sMAPE is below the
$5$th percentile of the random-pool distribution at every cell;
the Qwen row uses the L0\_100 Qwen-Scope variant of
Section~\ref{sec:models-saes}.}
\label{tab:resample}
\end{table*}

The original fixed-seed random pool included zero-firing features
whose sMAPE is artificially zero by the $\varepsilon$ denominator
floor (Section~\ref{sec:mech-invariance}); under proper
magnitude-matched resampling, the random distribution is shifted
upward and the medical-random gap is consequently larger than
the fixed-seed analysis suggested.

\section{Behavioral decomposition and option shuffling}

\subsection{Per-case NL-NF gap decomposition}
\label{app:gap_decomposition}

Table~\ref{tab:decomp_4b}--\ref{tab:decomp_qwen} report the
per-case decomposition supporting the adjacent-miscalibration
framing of Section~\ref{sec:deferred_class}. Adjacency is computed
on the gold acuity scale ${A\to B\to C\to D}$ as integer distance
$|\text{NL letter} - \text{NF letter}|$; ``adjacent'' means
distance $=1$.

\begin{table*}[ht]
\centering
\footnotesize
\setlength{\tabcolsep}{3pt}
\begin{tabular}{lccccc}
\toprule
Case & gold & NL & NF (both) & adj.\ & 5-way \\
\midrule
\multicolumn{6}{l}{\textbf{4B NF-only-right ($n=14$, gap-driver)}} \\
E3 & C & B & C & $\checkmark$ & --- \\
E4 & C & B & C & $\checkmark$ & --- \\
E10 & C/D & B & C & $\checkmark$ & --- \\
E11 & C/D & B & C & $\checkmark$ & --- \\
E22 & C/D & B & C & $\checkmark$ & --- \\
E25 & C & B & C & $\checkmark$ & --- \\
F1 & C/D & B & C & $\checkmark$ & --- \\
F3 & C & B & C & $\checkmark$ & --- \\
F4 & C & B & C & $\checkmark$ & --- \\
F10 & C/D & B & C & $\checkmark$ & --- \\
F19 & B & A & B & $\checkmark$ & --- \\
MH1 & C & B & C & $\checkmark$ & --- \\
NH1 & C & B & C & $\checkmark$ & --- \\
NH3 & C & B & C & $\checkmark$ & --- \\
\bottomrule
\end{tabular}
\caption{Per-case decomposition of 4B's NL${>}$NF
NF-only-right stratum ($n=14$). All $14$ are
single-acuity-step under-triages in NL; $13$ follow the exact
``NL${=}$B, gold${=}$C, NF${\to}$C'' pattern. No unanimous
\textsc{deferred} cases.}
\label{tab:decomp_4b}
\end{table*}

\begin{table}[ht]
\centering
\footnotesize
\setlength{\tabcolsep}{3pt}
\begin{tabular}{lccccc}
\toprule
Case & gold & NL & NF (both) & adj.\ & 5-way \\
\midrule
\multicolumn{6}{l}{\textbf{12B NL-only-right ($n=6$, gap-driver)}} \\
E19 & B & B & C & $\checkmark$ & --- \\
F3 & C & C & B & $\checkmark$ & --- \\
F7 & C/D & C & A & $\times$ & --- \\
F11 & C/D & C & B & $\checkmark$ & --- \\
F13 & D & D & C & $\checkmark$ & --- \\
NH3 & C & C & D & $\checkmark$ & --- \\
\midrule
\multicolumn{6}{l}{\textbf{12B unanimous \textsc{deferred} (all in both-right)}} \\
F15 & C/D & C & B & --- & \textsc{def} \\
F19 & B   & B & B & --- & \textsc{def} \\
F23 & A/B & B & B & --- & \textsc{def} \\
F24 & B   & B & B & --- & \textsc{def} \\
\bottomrule
\end{tabular}
\caption{12B's NL${>}$NF NL-only-right stratum ($n=6$,
$5$ adjacent miscalibrations) and the four unanimous
\textsc{deferred} cases. The deferrals flatten to gold-compatible
letters and live in both-right, contributing zero to the
gap.}
\label{tab:decomp_12b}
\end{table}

\begin{table}[ht]
\centering
\footnotesize
\setlength{\tabcolsep}{3pt}
\begin{tabular}{lccccc}
\toprule
Case & gold & NL & NF (both) & adj.\ & 5-way \\
\midrule
\multicolumn{6}{l}{\textbf{Qwen NL-only-right ($n=8$)}} \\
E3 & C & C & B & $\checkmark$ & --- \\
E4 & C & C & B & $\checkmark$ & --- \\
E8 & A & A & B & $\checkmark$ & --- \\
E17 & A & A & B & $\checkmark$ & --- \\
F1 & C/D & C & B & $\checkmark$ & --- \\
F4 & C & C & B & $\checkmark$ & --- \\
F11 & C/D & C & B & $\checkmark$ & --- \\
F25 & C & C & D & $\checkmark$ & --- \\
\midrule
\multicolumn{6}{l}{\textbf{Qwen NF-only-right ($n=6$)}} \\
E6 & B/C & A & C & $\times$ & --- \\
E7 & C/D & A & C & $\times$ & --- \\
F6 & B/C & A & C & $\times$ & --- \\
F7 & C/D & A & C & $\times$ & --- \\
F10 & C/D & A & C & $\times$ & --- \\
F24 & B & A & B & $\checkmark$ & --- \\
\bottomrule
\end{tabular}
\caption{Qwen3-8B has both strata at lower magnitude. The
NL-only-right stratum is $8/8$ adjacent miscalibrations.
The NF-only-right stratum exhibits a distinctive 2-step
NL${=}$A under-triage pattern on $5/6$ cases.}
\label{tab:decomp_qwen}
\end{table}

\subsection{Option-order shuffle on 4B}
\label{app:option_shuffle}

The Section~\ref{sec:deferred_class} ``NL${=}$B, gold${=}$C''
pattern on 4B admits two simple alternative explanations. The
\textbf{position-2 bias} hypothesis: the model prefers whatever
letter sits in position-$2$ in the canonical A--D layout. The
\textbf{content prior} hypothesis: the model prefers a specific
clinical disposition independent of which letter holds it.

We disambiguate by shuffling the A--D label assignments to the
same option texts. For each of the $60$ NL prompts and each model,
we run all $23$ non-identity permutations of $\{A,B,C,D\}$. Each
permutation rewrites the prompt with shuffled letter assignments
while keeping the option texts verbatim --- so, e.g., ``See a
doctor within 24--48 hours'' may sit under letter $A$, $B$, $C$,
or $D$ depending on the shuffle. We re-run forced-letter generation
on each shuffled prompt under greedy decoding ($60 \times 23 =
1380$ NL forward passes per model). We then ask: how often does
the model pick the same \emph{letter} as in canonical, and how
often does it pick the same \emph{content}?

\begin{table*}[ht]
\centering
\footnotesize
\setlength{\tabcolsep}{3pt}
\begin{tabular}{lccc}
\toprule
Signal & 4B & 12B & Qwen \\
\midrule
\multicolumn{4}{l}{\textbf{K=23 exhaustive (case-clustered 95\% CI)}} \\
Same-letter (chance $25\%$)  & $22.4$ \tiny{$[21.4,23.4]$} & $20.8$ \tiny{$[18.6,23.0]$} & $23.3$ \tiny{$[20.5,26.5]$} \\
Same-content (chance $25\%$) & $64.5$ \tiny{$[55.9,73.0]$} & $80.3$ \tiny{$[73.6,86.6]$} & $82.6$ \tiny{$[76.2,88.4]$} \\
Shuffled NL acc              & $69.8$ \tiny{$[60.7,78.3]$} & $76.3$ \tiny{$[66.3,85.3]$} & $75.4$ \tiny{$[66.0,84.5]$} \\
\midrule
Canonical NL acc             & $55.0\%$ & $81.7\%$ & $75.0\%$ \\
NF (free-text) acc           & $71.7\%$ & $71.7\%$ & $68.3\%$ \\
Shuffled $-$ NF (pp)         & $-1.9$ (ns)& $+4.6$  & $+7.1$  \\
\midrule
Picks ``ER now'' content     & $\mathbf{2/1380}$ & $126/1380$ & $97/1380$ \\
                             & $(0.14\%)$ & $(9.1\%)$ & $(7.0\%)$ \\
\bottomrule
\end{tabular}
\caption{Exhaustive option-order shuffle ($60$ cases $\times$ all
$23$ non-identity permutations of A--D label assignments per case,
per model, $4140$ forward passes total). The same-letter rates'
case-clustered $95\%$ CIs exclude the $25\%$ chance baseline at
4B and 12B and contain it at Qwen, ruling out position bias at
every scale. The same-content rates are far above chance at every
model, isolating a strong content prior. At 4B shuffled
forced-letter accuracy ($69.8\%$) is \emph{statistically
indistinguishable} from NF accuracy ($71.7\%$): the entire
NL-NF gap at 4B is consistent with a canonical
answer-key-mapping artifact. At 12B and Qwen, the canonical
mapping is approximately neutral but shuffled NL still beats NF
by $\sim 5$--$7$pp, indicating a separate NF-mode
adjacent-miscalibration penalty (Section~\ref{sec:deferred_class})
that label randomization does not remove. Capability scaling
shows up in the ``ER now'' row: 4B almost never recommends ER
even when the label is favorable ($0.14\%$), while 12B and Qwen
do so $7$--$9\%$ of the time. (For comparison, the K=3 random
subsample reported in an earlier draft happened to yield
exactly $71.7\%$ shuffled NL at 4B; the exhaustive K=23 estimate
of $69.8\%$ is the precise value with a $95\%$ CI containing NF.)}
\label{tab:option_shuffle}
\end{table*}

The cross-model picture cleanly separates two distinct sources of
format-dependent accuracy.

\textbf{(a) Canonical answer-key mapping $\times$ content-prior
interaction.} This source is scale-dependent. At 4B, the canonical
A--D layout misroutes the model's ``24--48h care'' content prior
to a wrong letter on the gold-$C$ cases, costing
$\sim 17$pp. At 12B and Qwen, the canonical mapping is
approximately neutral.

\textbf{(b) NF-mode adjacent miscalibration.} This source is
present at 12B and Qwen but absent at 4B, where shuffled NL is
statistically indistinguishable from NF.

Across all three models, position is not a factor (same-letter
rates near chance, two of three CIs below chance) and content
priors are strong (same-content rates $64$--$82\%$). The canonical
NL accuracy spread of $55$--$82$pp narrows to $\sim 70$--$76\%$
shuffled once the canonical answer-key binding is removed.

\section{Decision-token validations}

\subsection{Decision-token logit attribution: full results}
\label{app:logit_attribution}

The per-feature linear projection is defined in
Section~\ref{sec:methods_decision_token}. For every active SAE feature
and letter, we compute Eq.~\ref{eq:contrib}. Identified medical
features retain their fixed label; each remaining feature is classified
per case as scaffold-peaking when its maximum activation after the
shared vignette exceeds its vignette maximum by more than $1\%$, and
as vignette-peaking otherwise. This assigns every active feature by
its observed peak location rather than leaving an uninterpreted
remainder category.

\begin{table*}[ht]
\centering
\scriptsize
\setlength{\tabcolsep}{3.5pt}
\begin{tabular}{llccc}
\toprule
Model & Category & Unsigned share & Positive-letter share & Signed margin share \\
& & mean [95\% CI] & mean [95\% CI] & mean [95\% CI] \\
\midrule
\multirow{3}{*}{4B L29}
  & medical (3 features) & $0.0\%$ & $0.0\%$ & $0.0\%$ \\
  & scaffold-peaking & $\mathbf{93.2\%}$ \tiny{[91.8, 94.4]} & $\mathbf{96.1\%}$ \tiny{[95.2, 96.9]} & $115.0\%$ \tiny{[94.7, 152.4]} \\
  & vignette-peaking & $6.8\%$ \tiny{[5.6, 8.2]} & $3.9\%$ \tiny{[3.1, 4.9]} & $-15.0\%$ \tiny{[-52.4, 5.2]} \\
\midrule
\multirow{3}{*}{12B L31}
  & medical (3 features) & $0.0\%$ & $0.0\%$ & $0.0\%$ \\
  & scaffold-peaking & $\mathbf{91.9\%}$ \tiny{[90.9, 92.9]} & $\mathbf{96.0\%}$ \tiny{[94.9, 97.1]} & $\mathbf{77.4\%}$ \tiny{[72.5, 81.3]} \\
  & vignette-peaking & $8.1\%$ \tiny{[7.1, 9.1]} & $4.0\%$ \tiny{[2.9, 5.2]} & $22.6\%$ \tiny{[18.7, 27.5]} \\
\bottomrule
\end{tabular}
\caption{Exact SAE-feature linear projection at the NL decision
token ($n{=}60$; case-bootstrap $95\%$ CIs, $10{,}000$ resamples).
Unsigned share sums $|\mathrm{contrib}(f,\ell)|$ over all active
features and A--D letters without within-category cancellation;
positive-letter share uses positive contributions to the predicted
letter. Signed margin share is for predicted versus runner-up letter
and may exceed $100\%$ or be negative when categories oppose each
other. The projection excludes later layers, final normalization,
and SAE reconstruction error and is not a causal effect.}
\label{tab:logit_attribution}
\end{table*}

\paragraph{Feature and category checks.}
The contrastively-identified medical features are
$\{12570, 893, 12845\}$ at 4B L29 and $\{130, 85, 4773\}$ at 12B
L31. All remaining JumpReLU-active features at the decision token
are assigned by the peak-location rule above. Reproduction checks
against the stored signed category-by-letter sums are included with
the released analysis output.

\paragraph{Activation diagnostic.} Before attributing the medical
features' zero contribution to orthogonality with the letter
directions, we verified that the medical features are not active
at this position at all. At 4B L29, the three medical features
have zero activation at the NL pre-generation token in
$60/60$ cases; at 12B L31, the analogous three features have
zero activation in $60/60$ cases. The features fire on
clinical-narrative tokens during clinical reading
(Appendix~\ref{app:per_token}) and then go silent at the
decision token. The zero linear contribution in
Table~\ref{tab:logit_attribution} therefore reflects a categorical
absence of medical-feature activity at the decision point, not a
near-orthogonality between active medical features and the
unembedding directions.

\subsection{Decision-token top-feature characterization}
\label{app:decision_token_features}

The exact logit attribution
(Appendix~\ref{app:logit_attribution}) includes every active feature
and groups it by peak location. A complementary question is which
features are most active at the NL decision token in the first
place, regardless of direction-analysis selection, and where
those features peak on B-prompt tokens.

\paragraph{Method.} For each case at each model, take the top-$20$
features by activation at the NL decision token and at the NF
decision token. Compute Jaccard overlap $|NL_{20} \cap NF_{20}| /
|NL_{20} \cup NF_{20}|$. For features in $NL_{20} \setminus NF_{20}$,
record where the feature reaches its peak activation on the B
(NL) prompt: ``scaffold'' if the peak token sits outside the
shared clinical vignette (i.e.\ in the scaffold instructions or
A--D answer-key block), ``vignette'' otherwise. Symmetric
analysis for $NF_{20} \setminus NL_{20}$ on the D (NF) prompt.

\begin{table*}[ht]
\centering
\footnotesize
\setlength{\tabcolsep}{3pt}
\begin{tabular}{lccc}
\toprule
Model & NL$\cap$NF top-$20$ & NL-only peak & NF-only peak \\
      & (Jaccard)           & on B scaffold& on D vignette \\
\midrule
4B L29   & $0.000$ $[0.00, 0.00]$ & $87.0\%$ (med.\ $90\%$) & $27.8\%$ \\
12B L31  & $0.001$ $[0.00, 0.00]$ & $88.3\%$ (med.\ $90\%$) & $\phantom{0}8.9\%$ \\
Qwen L31 & $0.324$ $[0.25, 0.38]$ & $94.7\%$ (med.\ $100\%$)& $10.4\%$ \\
\bottomrule
\end{tabular}
\caption{Decision-token top-$20$ feature characterization. At
Gemma the NL and NF top-$20$ sets are disjoint
(Jaccard $\approx 0$); at Qwen they share about a third. Across
all three models, $87$--$95\%$ of features unique to NL's top-$20$
peak on B-prompt scaffold tokens outside the shared vignette ---
the direct quantification of ``scaffold-primary at NL.'' Features
unique to NF's top-$20$ peak on D-prompt vignette tokens only
$8$--$28\%$ of the time, indicating that ``content-primary at NF''
is weaker than the NL-side scaffold finding; many NF-only
top-$20$ features peak elsewhere on the D prompt (e.g.\ closing
punctuation, question-mark tokens).}
\label{tab:decision_token_features_full}
\end{table*}

\paragraph{None of the identified medical features are in either top-$20$
at any model.} Across all $60 \times 3 = 180$ (case, model) pairs,
the count of medical features appearing in $NL_{20}$ is $0$ and in
$NF_{20}$ is $0$. This corroborates the logit-attribution
finding: medical-content detectors are not in the top of the
active feature pool at the decision token; they are not active at
that position at all.

\paragraph{Reading the Qwen-vs-Gemma asymmetry.}
The Qwen Jaccard of $0.32$ (vs.\ $\approx 0$ at Gemma) shows that
Qwen shares some decision-token features across formats. The
``$94.7\%$ scaffold peak'' asymmetry on NL-only features
nonetheless holds at Qwen; the scaffold-primary pattern is
robust across the Jaccard difference.

\section{Input-style robustness}

\subsection{SL--SF mechanistic invariance}
\label{app:input-style-robustness}

This appendix carries the full numbers behind the input-style
robustness paragraph in \S\ref{sec:phase1b_robustness}. The
finding: medical features are significantly more format-invariant
than magnitude-matched random features under the structured-input
SL-SF pair too, not only the natural-input NL-NF pair.

\paragraph{Protocol.} The SL-SF protocol mirrors the NL-NF
mechanistic invariance test of
Section~\ref{sec:mech-invariance}: forward-pass each model under
the SL and SF prompts at the headline layer (4B L29, 12B L31,
Qwen L31), max-pool feature activations over user-content tokens,
compute medical-random sMAPE on the $3$ contrastively-identified
medical features versus $30$ magnitude-matched random features, and
bootstrap paired $95\%$ CIs ($2{,}000$ resamples).

\begin{table*}[ht]
\centering
\footnotesize
\setlength{\tabcolsep}{4pt}
\begin{tabular}{lcccc}
\toprule
Model / Layer & med sMAPE & rnd sMAPE & paired $\Delta$ [95\% CI] & Sig \\
\midrule
4B L29  & $0.0042$ & $0.0827$ & $-0.0810$ $[-0.096, -0.066]$ & $\ast$ \\
12B L31 & $0.0025$ & $0.0692$ & $-0.0618$ $[-0.083, -0.034]$ & $\ast$ \\
Qwen L31 & $0.0020$ & $0.1126$ & $\mathbf{-0.1501}$ $[-0.183, -0.117]$ & $\ast$ \\
\bottomrule
\end{tabular}
\caption{SL-SF mechanistic invariance at the headline layer per
model. $\Delta=$ medical${-}$random sMAPE; lower $=$ more
invariant. $\ast$ marks cells whose bootstrap $95\%$ CI excludes
zero. The vignette-mask check gives shared-content sMAPE
$\approx 0.002$--$0.006$ for both pools across all three models,
as expected from causal masking.}
\label{tab:sl_sf_invariance}
\end{table*}

\paragraph{Reproducibility.}
The per-model SL--SF arrays are included in the released results.

\section{Supplementary Layer-by-Layer Probing Results}
\label{app:supplementary layer-by-layer probing results}
This appendix provides the complete, layer-by-layer experimental record for all evaluated prompt format transitions. It includes the comprehensive matrix of empirical $p$-values for statistical significance testing, followed by the extended ROC-AUC and PR-AUC performance trajectories for all directional pathways. 

\begin{figure}[h]
    \centering
    \includegraphics[width=\columnwidth]{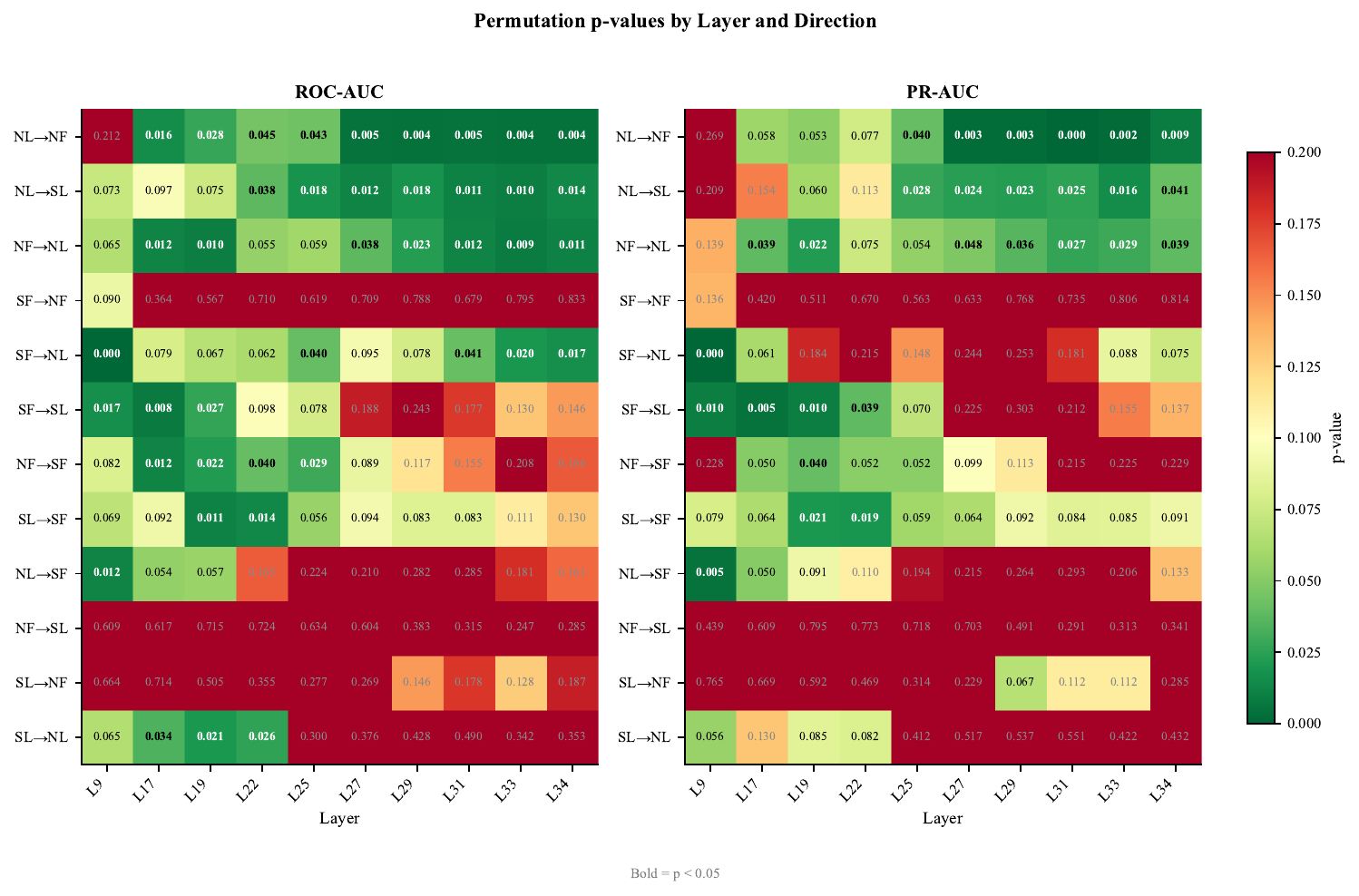}
    \caption{Comprehensive permutation $p$-values heatmap for all directions and layers (original setup).}
    \label{fig:p_values_matrix}
\end{figure}

\begin{figure}[h]
    \centering
    \includegraphics[width=\columnwidth]{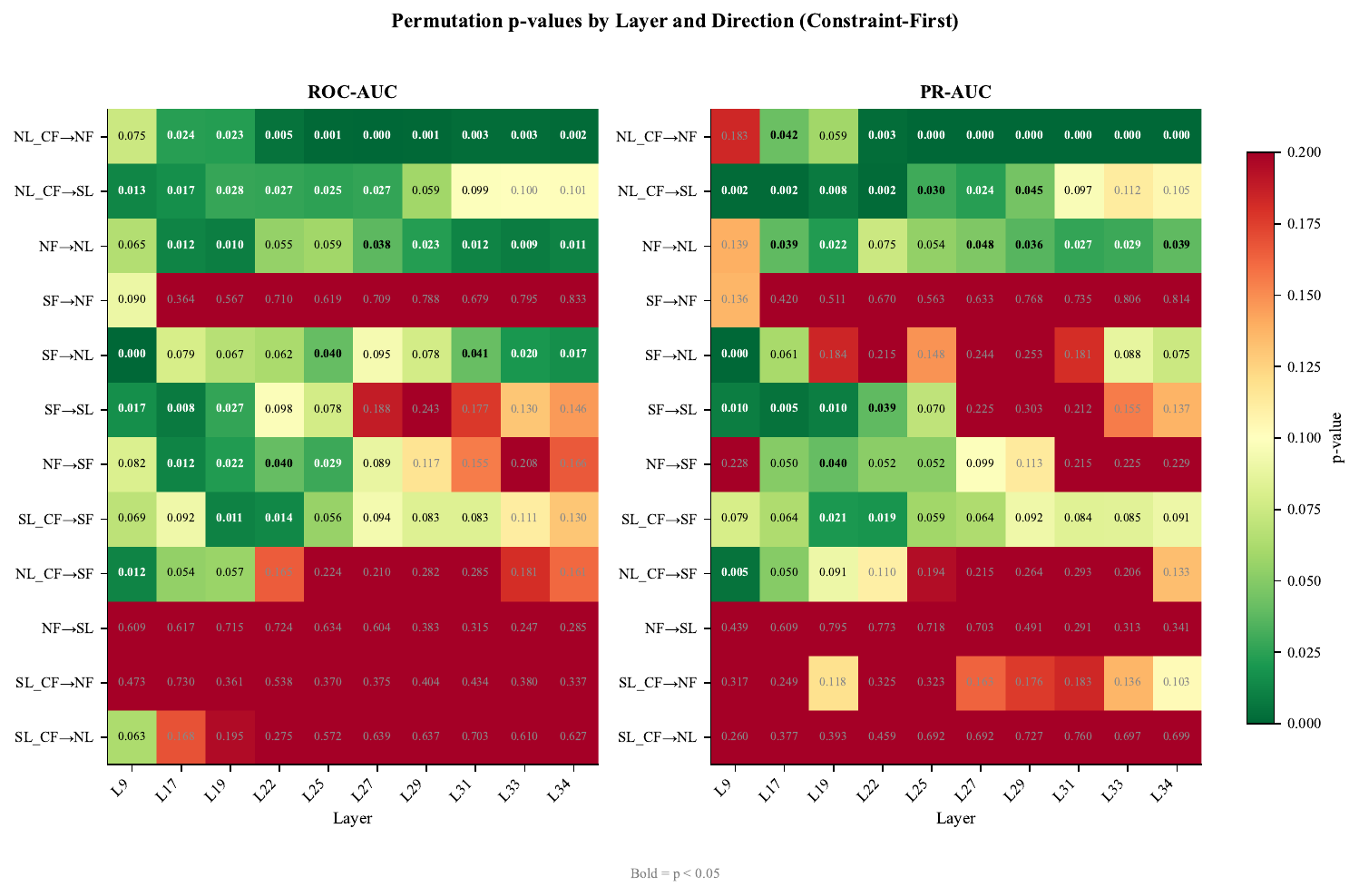}
    \caption{Comprehensive permutation $p$-values heatmap for all directions and layers (constraint-first setup).}
    \label{fig:p_values_matrix_cf}
\end{figure}

\begin{figure}[h]
    \centering
    \includegraphics[width=\columnwidth]{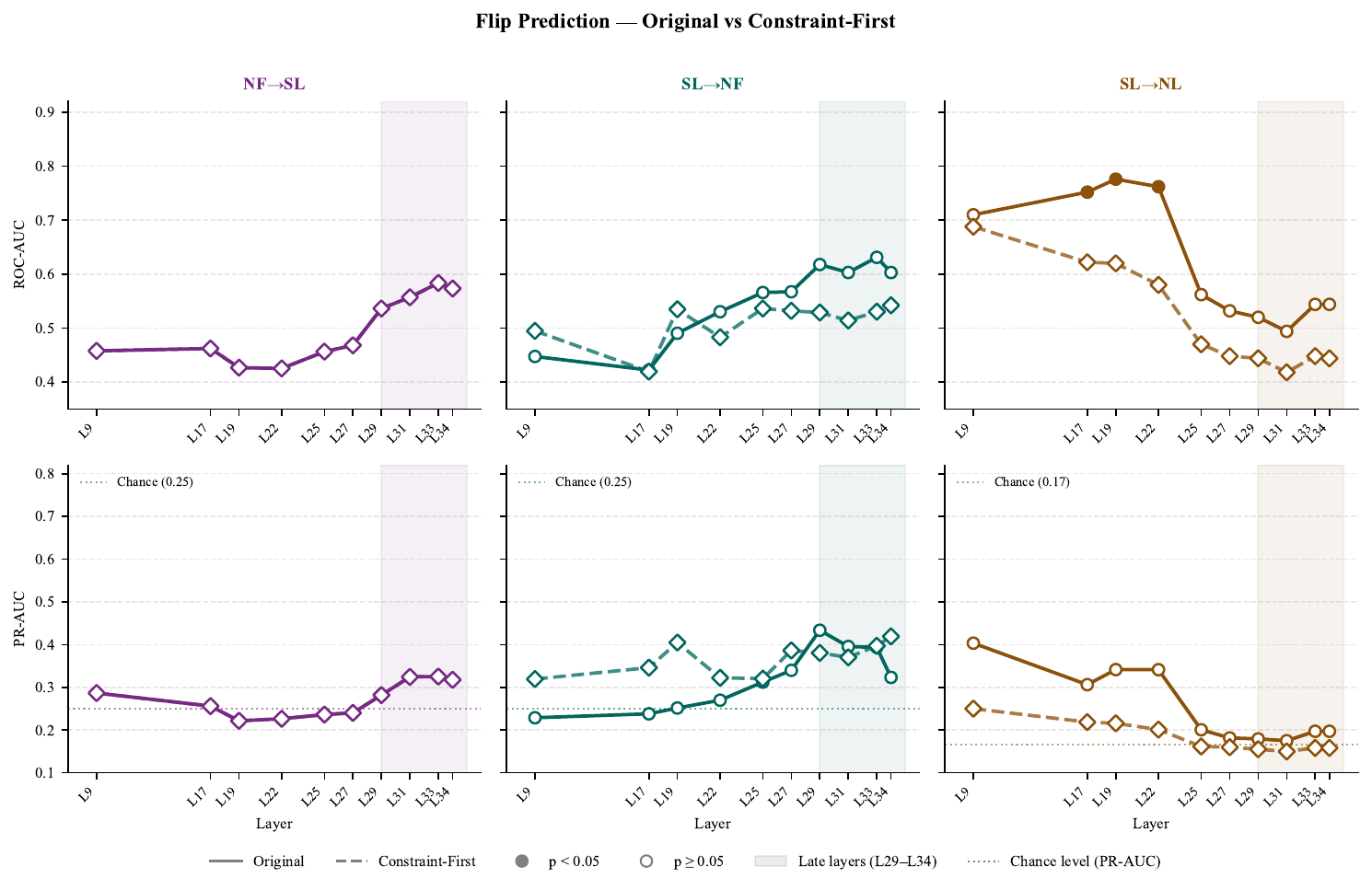}
    \caption{$\text{NF}\rightarrow \text{SL}$, $\text{SL}\rightarrow \text{NF}$, $\text{SL}\rightarrow \text{NL}$ performances}
    \label{fig:performance_app1}
\end{figure}

\begin{figure}[h]
    \centering
    \includegraphics[width=\columnwidth]{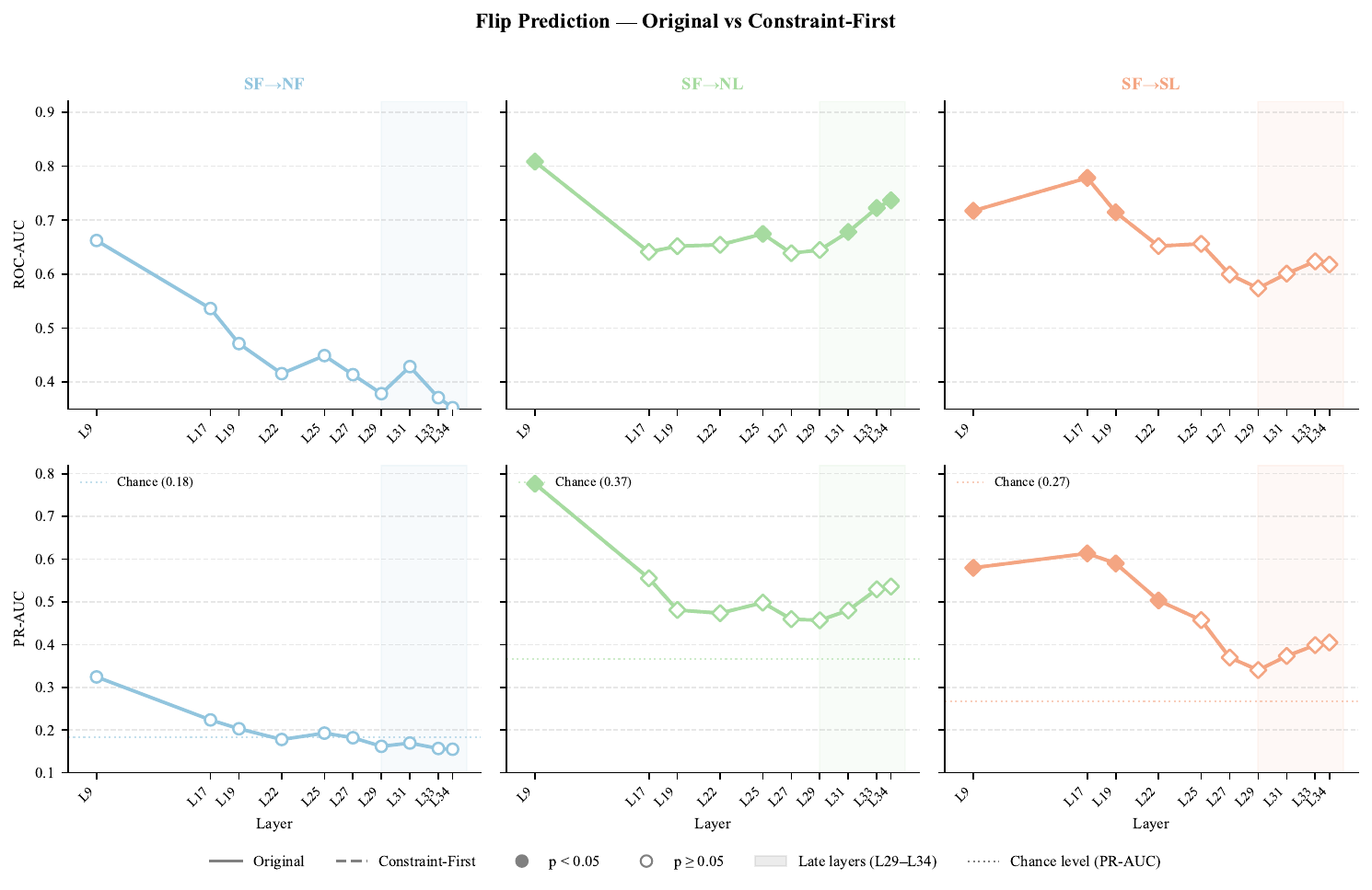}
    \caption{$\text{SF}\rightarrow \text{NF}$, $\text{SF}\rightarrow \text{NL}$, $\text{SF}\rightarrow \text{SL}$ performances}
    \label{fig:performance_app2}
\end{figure}

\begin{figure}[h]
    \centering
    \includegraphics[width=\columnwidth]{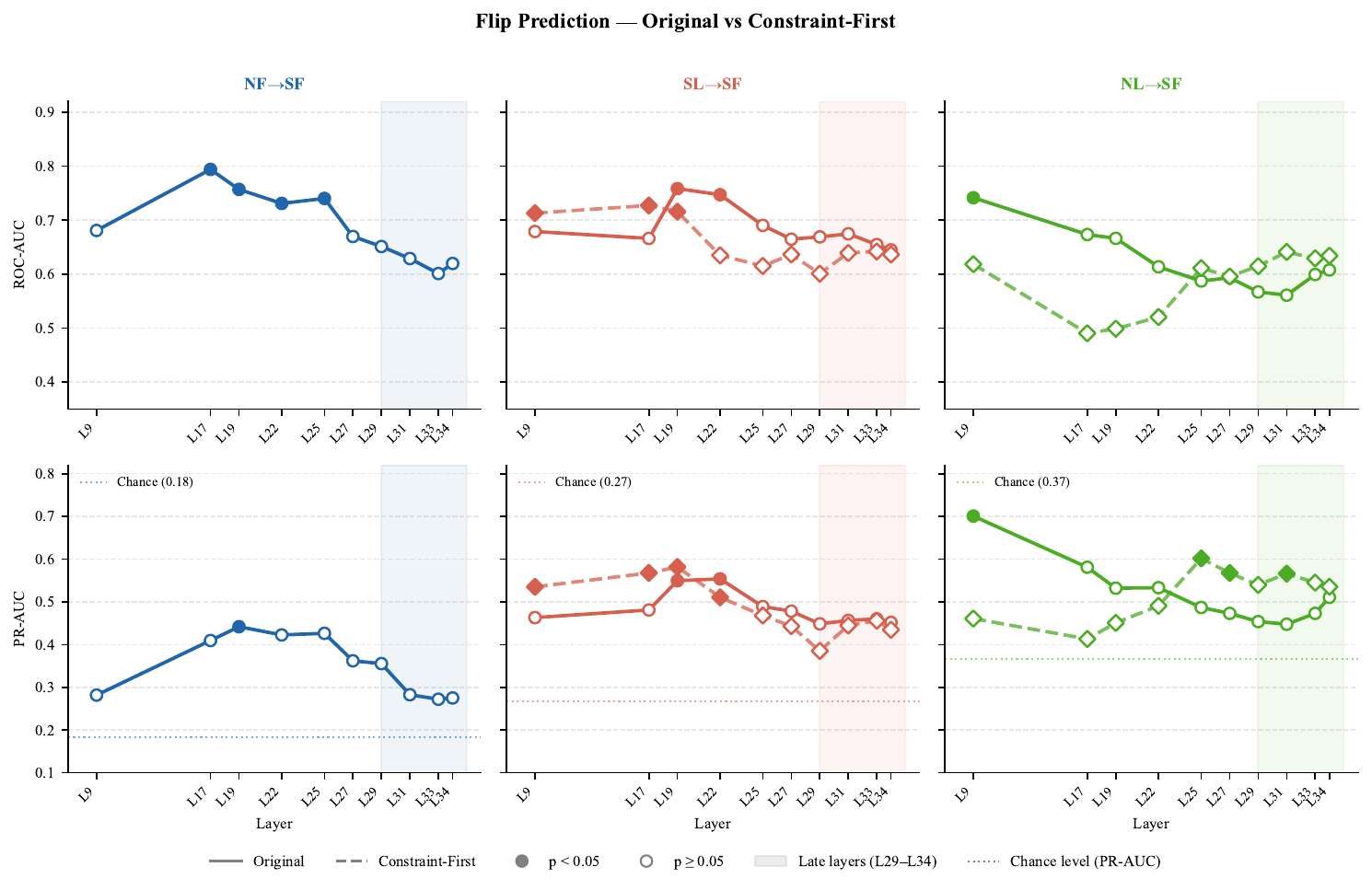}
    \caption{$\text{NF}\rightarrow \text{SF}$, $\text{SL}\rightarrow \text{SF}$, $\text{NL}\rightarrow \text{SF}$ performances}
    \label{fig:performance_app3}
\end{figure}

\end{document}